\documentclass{article} 
\usepackage{iclr2026_conference,times}

\usepackage{amsmath,amsfonts,bm}

\def\eqref#1{equation~\ref{#1}}

\def\1{\bm{1}}

\DeclareMathAlphabet{\mathsfit}{\encodingdefault}{\sfdefault}{m}{sl}
\SetMathAlphabet{\mathsfit}{bold}{\encodingdefault}{\sfdefault}{bx}{n}

\usepackage{hyperref}
\usepackage{url}
\usepackage{graphicx}
\usepackage{cleveref}
\crefname{figure}{Fig.}{Figs.}
\usepackage{enumitem} 
\usepackage{booktabs}
\usepackage{subcaption}
\usepackage{caption}
\usepackage{multirow}

\usepackage{floatrow}
\newfloatcommand{capbtabbox}{table}[][\FBwidth]

\title{Optimizing ID Consistency in Multimodal Large Models:  Facial Restoration via Alignment, Entanglement, and Disentanglement}

\author{Yuran Dong, Hang Dai$^{*}$, Mang Ye\thanks{Corresponding authors.} \\
   National Engineering Research Center for Multimedia Software \\                   
  School of Computer Science, Wuhan University, Wuhan, China \\ 
\texttt{\{dongyuran, daihang, yemang\}@whu.edu.cn} \\
}

\iclrfinalcopy 
\begin{document}

\maketitle

\begin{figure*}[h]
  \centering
  \includegraphics[width=1\linewidth]{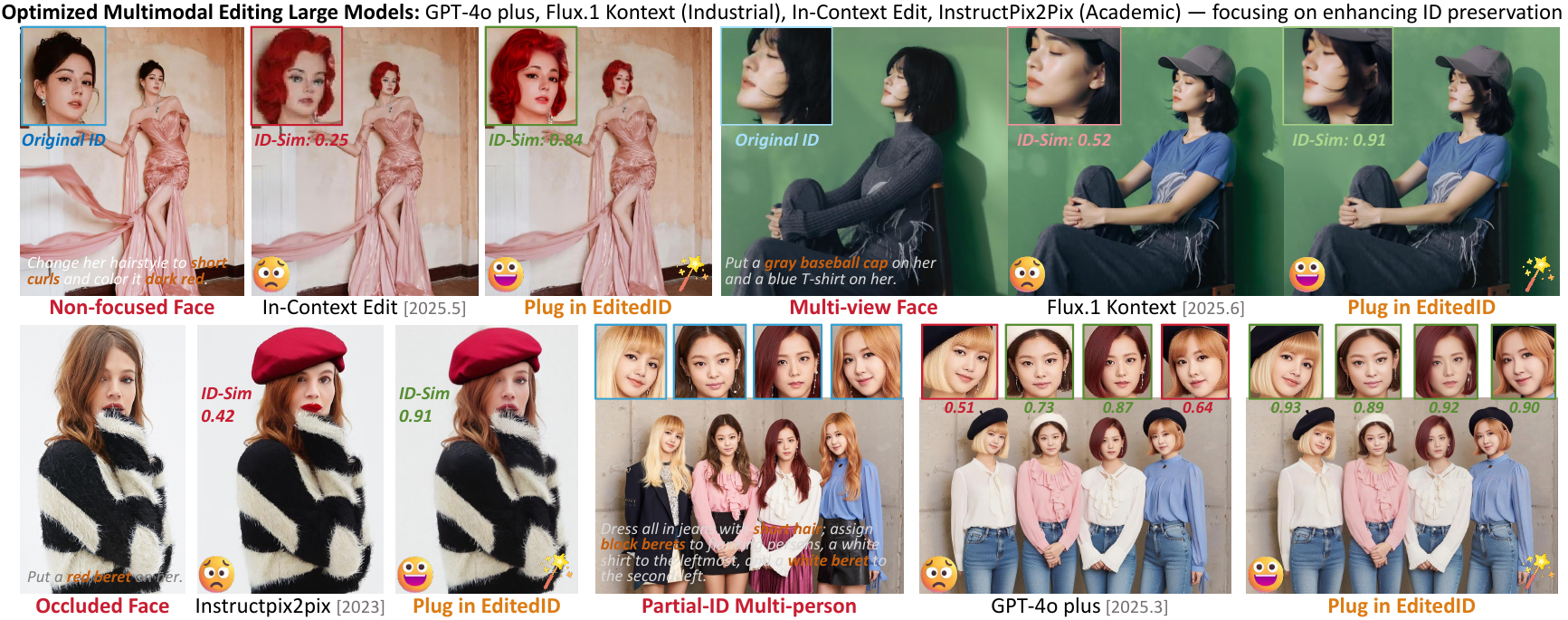}
    \caption{EditedID Optimized Multimodal Editing Large Models (Academic/Industrial): ID consistency capability. ID-Sim (same ID $>0.7$, green border; otherwise red) indicates identity preservation in challenging scenarios: Non-focused, Multi-view, Occluded Faces, Partial-ID Multi-persons.}
  \label{fig:abstract}
\end{figure*}

\begin{abstract}
Multimodal editing large models have demonstrated powerful editing capabilities across diverse tasks. However, a persistent and long-standing limitation is the decline in facial identity (ID) consistency during realistic portrait editing. Due to the human eye’s high sensitivity to facial features, such inconsistency significantly hinders the practical deployment of these models. 
Current facial ID preservation methods struggle to achieve consistent restoration of both facial identity and edited element IP due to \emph{Cross-source Distribution Bias} and \emph{Cross-source Feature Contamination}.
To address these issues, we propose \textbf{EditedID}, an Alignment-Disentanglement-Entanglement framework for robust identity-specific facial restoration. By systematically analyzing diffusion trajectories, sampler behaviors, and attention properties, we introduce three key components: 1) Adaptive mixing strategy that aligns cross-source latent representations throughout the diffusion process. 2) Hybrid solver that disentangles source-specific identity attributes and details. 3) Attentional gating mechanism that selectively entangles visual elements. Extensive experiments show that EditedID achieves state-of-the-art performance in preserving original facial ID and edited element IP consistency.
As a training-free and plug-and-play solution, it establishes a new benchmark for practical and reliable single/multi-person facial identity restoration in open-world settings, paving the way for the deployment of multimodal editing large models in real-person editing scenarios.
The code is available at \url{https://github.com/NDYBSNDY/EditedID}. 
\end{abstract}

\section{Introduction}\label{S1}
Recent years have witnessed growing interest in Multimodal Editing Large Models \cite{zhang2025context, Liu2025Step1XEditAP, flux, gpt4o,deng2025emerging, wu2025qwen} owing to their broad applicability and practical value. These models function by interpreting user instructions to enable image editing.
Although these models are effective for identity preservation in cartoons, they degrade significantly with complex long-instruction real-person fashion editing. Detailed prompts like ``make him wear a light gray jacket with black-framed glasses" frequently induce facial artifacts. 
Academic models, such as In-ContextEdit \cite{zhang2025context} and InstructPix2Pix \cite{brooks2023instructpix2pix}, suffer from limited fine-tuning data, affecting facial feature extraction and causing progressive deterioration of facial edits with long instructions. 
Industrial models, such as GPT-4o Plus \cite{gpt4o}, Qwen-Image-Edit \cite{wu2025qwen} and Flux.1 Kontext \cite{flux} prioritize LLM-driven textual controllability but neglect facial geometric constraints, producing random facial identities. Given the high sensitivity of human perception to facial features, even slight deviations in identity can render the results unusable. However, due to the confidentiality of real-world facial datasets, task-specific fine-tuning is often impractical. 
Multimodal editing large models face the long-standing challenge of achieving reliable identity preservation when performing complex, long-instruction edits \cite{dong2025pose,Wang2024TexFitTF} in real-person scenarios.

\begin{figure}[h]
  \centering
  \includegraphics[width=1\linewidth]{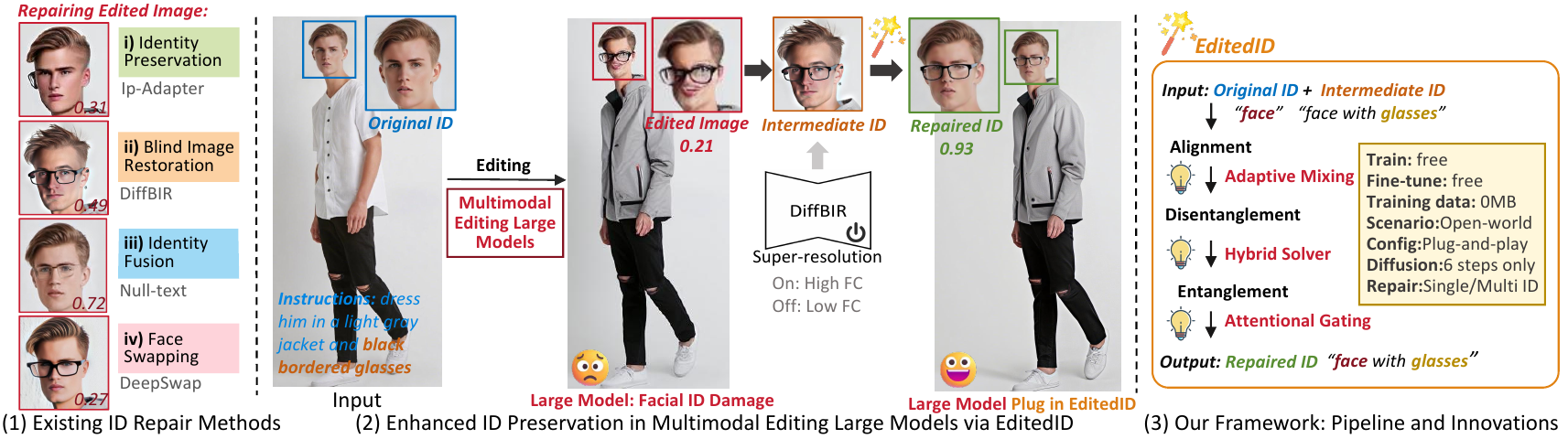}
    \caption{Enhanced by EditedID: ID preservation in Multimodal Editing Large Models. Comparison of repair effects and innovations. DiffBIR pre-repair triggered by Facial Corruption (FC) severity.}
  \label{fig:Problem}
\end{figure}

To optimize the ID consistency capability of existing multimodal editing large models, as shown in Fig. \ref{fig:Problem} (2), our goal is to achieve plug-and-play facial identity reconstruction. 
This involves restoring facial consistent with the Original ID while preserving edited Element IP (e.g., the color, size, or texture of accessories such as glasses) consistent with the Intermediate ID. 
In Fig. \ref{fig:Problem} (1), existing four types of facial consistency methods each exhibit limitations in achieving this objective:

\textbf{Cross-source Distribution Bias Leads to:}
\emph{i) Detail Loss in Identity Preservation.}
Identity-preserving approaches \cite{ye2025dreamid, Ohayon2024PosteriorMeanRF, Ye2023IPAdapterTC} often degrade photorealism by blurring fine facial details and introducing cartoon-like artifacts. This degradation primarily stems from Cross-source Distribution Bias. For instance, IP-Adapter \cite{Ye2023IPAdapterTC} merges coarse-grained identity features (learned from limited data) with high-resolution features from base diffusion models, leading to distribution mismatches and distortion artifacts.
\emph{ii) Original ID Lost in Blind Restoration.}
Generic facial restoration methods \cite{Hu2025UniversalIR, ito2025undertrained, lin2024diffbir} focus on facial super-resolution but neglect identity consistency. 
Although these models generate clear textures and plausible structures, Cross-source Distribution Bias between the generated facial features and the original distribution leads to random, non-ID-specific facial reconstruction.

\textbf{Cross-source Feature Contamination Leads to:}
\emph{iii) Element IP Loss in Identity Fusion.}
Fusion-based methods \cite{yao2025freegraftor, Nam2024DreamMatcherAM, 10328884, ye2023channel} act as a relaxed feature transfer mechanism. During cross-source feature (original facial and edited element) fusion, such methods often suffer from inter-feature contamination, leading to loss of fine-grained edited attributes. For instance, the ``black-framed" style of glasses is frequently lost, as these methods prioritize element-level consistency over accurate element attribute preservation.
\emph{iv) Edited Facial Noise in Face-Swapping.}
Face-swapping methods \cite{ye2025dreamid, wang2025dynamicface, Wang2024FaceSV, Liu2024,Li2024} exhibit high sensitivity to artifacts in edited facial. 
Effective under normal facial conditions, their performance degrades significantly in the presence of geometric or structural distortions, where Cross-source Feature Contamination (original ID and edited facial) and leads to loss of the original ID in swapping.
To address these limitations, we draw inspiration from 3D facial processing methods \cite{li2023megane}, which separate specific elements (e.g., glasses and faces) from cross-source objects and re-simulate physical interactions (e.g., lighting, position) to combine them into novel composites.
This inspires the core principle of our proposed \textbf{EditedID} (Fig. \ref{fig:Problem} (2) and (3)) in 2D scenarios: \textbf{Alignment}–\textbf{Disentanglement}–\textbf{Entanglement}.
EditedID is a diffusion-inversion-based ID-consistent facial reconstruction method. Specifically,
to mitigate \emph{Cross-source Distribution Bias}, we introduce Adaptive Mixing for dual-ID (Original and Intermediate ID) latent Alignment;
to isolate \emph{Cross-source Feature Contamination} in aligned identities and preserve personalized element features, we propose a Hybrid Solver for dual-ID latent Disentanglement.
Finally, via an Attentional Gating mechanism, we Entangle facial attributes from the Original ID with edited elements from the Intermediate ID.
Refer to Appendix \ref{Related Work} for related work.
Our contributions are as follows:
\begin{itemize} [left=-10pt, itemsep=0.001em, topsep=0.001em]
    \item[ ] \includegraphics[scale=0.05]{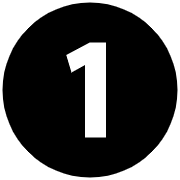} \textbf{Trajectory Insight — Adaptive Mixing:} By revealing the multi-solution and controllability of diffusion trajectories, we propose Adaptive Mixing—a cross-object feature fusion approach that mitigates \emph{Cross-source Distribution Bias} to avoid abrupt transitions and artifacts.
    \item[ ] \includegraphics[scale=0.05]{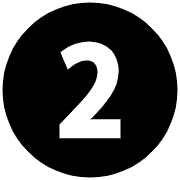} \textbf{Sampler Analysis – Hybrid Solver:} By leveraging DDIM identity retention and DPM-Solver++ detail enhancement properties, we design Hybrid Solver—a global-timestep hybrid sampling method that isolates \emph{Cross-source Feature Contamination} while preserving original identity-detail features.
    \item[ ] \includegraphics[scale=0.05]{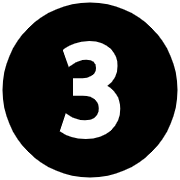} \textbf{Attention Mechanism – Attentional Gating:} 
    By uncovering the distinct roles of attention in diffusion processes, we introduce Attentional Gating—an specified element control mechanism that preserves single-element structures and balances multi-element interactions during entanglement.
    \item[ ] \includegraphics[scale=0.05]{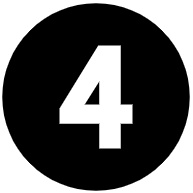} \textbf{Identity-Consistent Restoration Paradigm:} We propose a novel, training-free trajectory Alignment framework for real-world single- and multi-facial restoration, leveraging the Disentanglement and Entanglement of cross-source semantic-specific elements within pre-trained diffusion models. 
\end{itemize}


\section{Methodology}
\begin{figure}[h]
  \centering
  \includegraphics[width=1\linewidth]{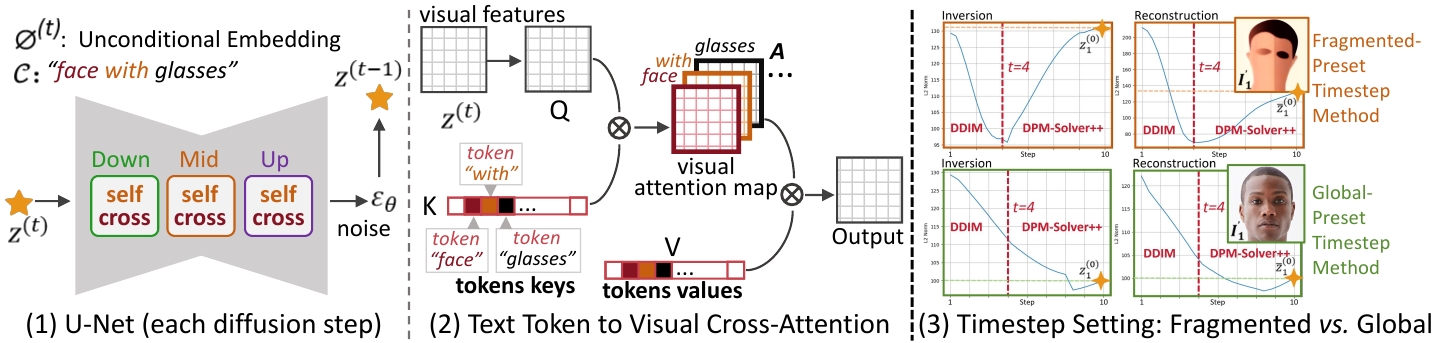}
  \caption{\textbf{Preliminaries:} (1) U-Net (2) cross-attention. \textbf{Hybrid solver:} (3) Timestep Init. (Sec.\ref{S3.2}).}
  \label{fig:Background}
\end{figure}
\textbf{Preliminaries.} Diffusion models synthesize an image $z^{(0)}$ by iteratively denoising random noise $z^{(T)}$ over $T$ steps. The inversion process diffuses a real image $z^{(0)}$ to noise $z^{(T)}$, while the reconstruction path reverses it via a denoising trajectory. 
Null-text optimization \cite{mokady2023null} bridges the deviation between reconstruction and inversion by introducing an unconditional embedding $\emptyset^{(t)}$, achieving accurate reconstruction trajectories of real images $\bar{z}^{(0)} \approx z^{(0)}$. This process is training-free.
At each step $t$, U-Net predicts noise $\varepsilon = \varepsilon_\theta(z^{(t)}, t)$ using $z^{(t)}$, conditional embedding $\mathcal{C}$, and $\emptyset^{(t)}$; the sampler then updates $z^{(t-1)}$ accordingly in \cref{fig:Background}(1).

\;\;\;\;\textbf{Diffusion Samplers:} DDPM \cite{ho2020denoising} adopts a Markov chain (e.g., $T=1000$), incurring high cost. DDIM \cite{Song2020DenoisingDI} accelerates sampling via non-Markovian steps (e.g., $T=50$). It reconstructs $\tilde{z}^{(0)}$ from $z^{(t)}$
and deterministically computes $z^{(k)}$, $k<t$. DPM-Solver++ \cite{Lu2022DPMSolverFS} treats reverse diffusion as an ODE, solved via high-order Taylor approximations. This enables rapid convergence with realistic details (details in Appendix \ref{Appendix A}).

\;\;\;\;\textbf{U-Net Attention:} U-Net includes cross-attention layers (\cref{fig:Background}(2)) that map text tokens to visual attention maps $A$. Query $Q$ is projected from image features $z^{(t)}$, while Key $K$ and Value $V$ from $\mathcal{C}$. Prompt-to-Prompt \cite{Hertz2022PrompttoPromptIE} manipulates attention maps for feature transfer. Null-text extends this to real-image editing via reconstruction trajectory.

\textbf{Problem Statement:} Given a source identity $I_1$ and an edited image $I_2$ (e.g., with black glasses), the goal is to reconstruct a new image $I_3$ that preserves $I_1$'s identity while retaining $I_2$'s semantics. For degraded $I_2$, we first apply DiffBIR \cite{lin2024diffbir} for super-resolution enhancement.

\subsection{Motivation}
\begin{figure*}[h]
  \centering
  \includegraphics[width=0.98\linewidth]{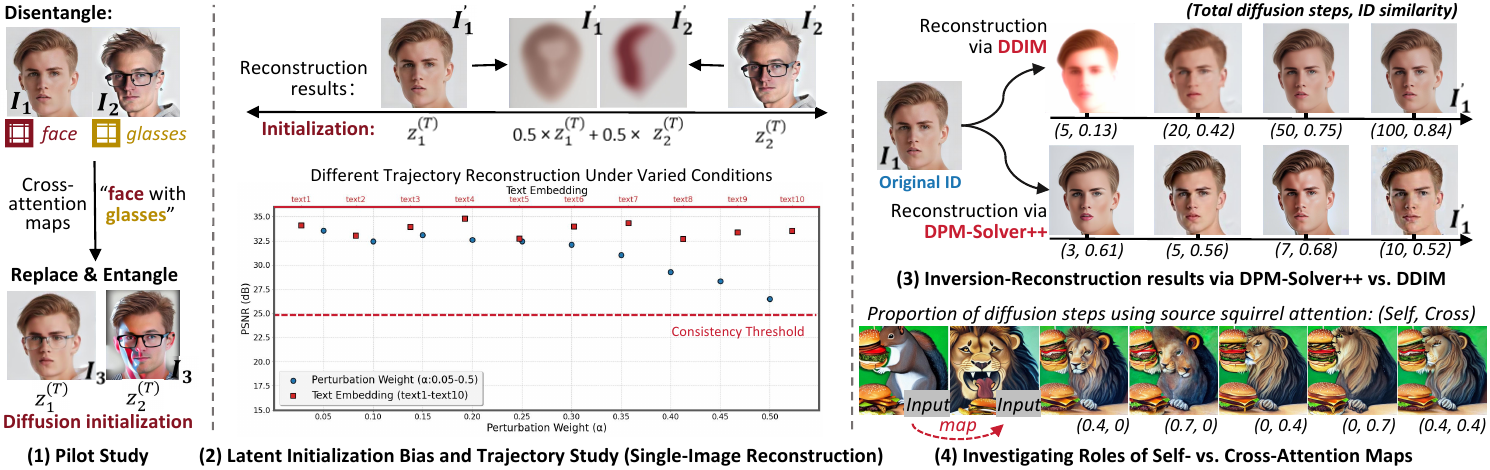}
    \caption{(1) Limitations of the Pilot Study and analysis of three contributing factors: (2) (3) (4).}
  \label{fig:Discoveries}
\end{figure*}
\noindent\textbf{Pilot Study.} 
We investigate Prompt-to-Prompt and Null-text for identity-consistent restoration of edited images, focusing on feature disentanglement and fusion via cross-attention.
In \cref{fig:Discoveries} (1), \emph{\textbf{Latent Initialization}} of $I_3$ is achieved by using either $z_1^{(T)}$ or $z_2^{(T)}$, which are obtained through \emph{\textbf{DDIM Sampler}} inversion of \emph{Dual-ID} ($I_1$ and $I_2$). During the diffusion process of $I_3$, \emph{\textbf{Cross-Attention}} maps ($I_1$→``face", $I_2$→``glasses") are fused to guide cross-source feature transfer.
However, $I_3$ exhibits asymmetric feature retention—starting from $z_1^{(T)}$ preserves ID (face) but loses edited attributes (glasses), and vice versa. 
Significant artifacts also emerge in $I_3$.
Key factors include:

\textbf{1) Latent Initialization Bias:} Different initial latents ($z_1^{(T)}$ \emph{vs.} $z_2^{(T)}$) produce incompatible attention maps, leading to poor fusion. Mixing latents ($0.5z_1^{(T)} + 0.5z_2^{(T)}$) amplifies artifacts due to diffusion nonlinearity (\cref{fig:Discoveries} (2)-top). 
To reduce cross-source latent space bias, we explore the controllability of diffusion trajectories. In \cref{fig:Discoveries} (2)-bottom, trajectories are perturbed via text embedding modifications or noise injection ($\alpha \in [0.05, 0.5]$). 
Higher PSNR \cite{Fardo2016AFE} indicates to greater reconstruction-original similarity ($\bar{z}^{(0)} \approx z^{(0)}$), consistency threshold: 25 dB.
Similar PSNR under different perturbations suggest identical reconstructions across trajectories. Our reveal key findings:

\;\;\;\;\textbf{Observation 1 — Trajectory:}
\textbf{i) Multi-solution.} Multiple trajectories can achieve same output (same PSNR) if fixed target $z^{(0)}$, demonstrating multi-path solvability ($\alpha=0.25$ or $text=text5$) of $z^{(0)}$.
\textbf{ii) Controllability.} $z^{(T)}$ and evolution jointly shape trajectories; controlled perturbations ($\alpha$) allow trajectory alteration without degrading fidelity (PSNR $>25$ dB).

\textbf{2) DDIM Sampler Loss Detail:} DDIM sampler introduces bias in detail fidelity, which is exacerbated by cross-source fusion and leads to artifacts. \cref{fig:Discoveries}(3) compares ID similarity \cite{deng2019arcface} ($>0.7$ same ID) between DPM-Solver++ and DDIM under different reconstruction steps.

\;\;\;\;\textbf{Observation 2 — Sampler:}
\textbf{i) DDIM: Identity over Details.} More diffusion steps ($>50$) with deterministic path preserves identity consistency (ID sim: 0.84), but detail loss due to 1st-order smoothing and error accumulation.
\textbf{ii) DPM-Solver++: Details over Identity.} Fewer diffusion steps ($<10$) with high-order Taylor expansion produce high-fidelity details, yet path deviation and prior interference cause identity loss (Avg. ID sim: 0.59). Detailed analysis in Appendix \ref{DPAnalysis}.



\textbf{3) Cross-Attention Weak Constraints:} Cross-source feature transfer via cross-attention alone yields weak constraints and ignores spatial coupling in self-attention. We analyze self \emph{vs.} cross roles by manipulating attention maps (``squirrel'' replace ``lion'') under varied strengths in \cref{fig:Discoveries} (4):

\;\;\;\;\textbf{Observation 3 — Attention:}
\textbf{i) Self: Single-Element Structure.} Self-attention encodes single-element structure; increasing self-strength enforces shape transfer (lion→squirrel).
\textbf{ii) Cross: Multi-Element Interaction.} Cross-attention encodes multi-element interaction; increasing cross-strength enhances semantic interactions (lion \emph{eating} hamburger).
\textbf{iii) Balance:} 
Coordinated cross/self-attention tuning is essential for structure-interaction trade-offs and coherence outcomes.

These observations motivate us to solve three core challenges, solved in Sec. \ref{S3.1}, Sec. \ref{S3.2}, Sec. \ref{S3.3}.
\begin{itemize}[left=0pt]
\item \textbf{C1:} \emph{How to align dual-ID diffusion trajectories to reduce cross-source latent initialization bias?}
\item \textbf{C2:} \emph{How to integrate DDIM and DPM-Solver++ for identity-detail preserving disentanglement?}
\item \textbf{C3:} \emph{How to coordinate cross/self-attention for structure-interaction balanced entanglement?}
\end{itemize}


\begin{figure*}[h]
  \centering
  \includegraphics[width=1\linewidth]{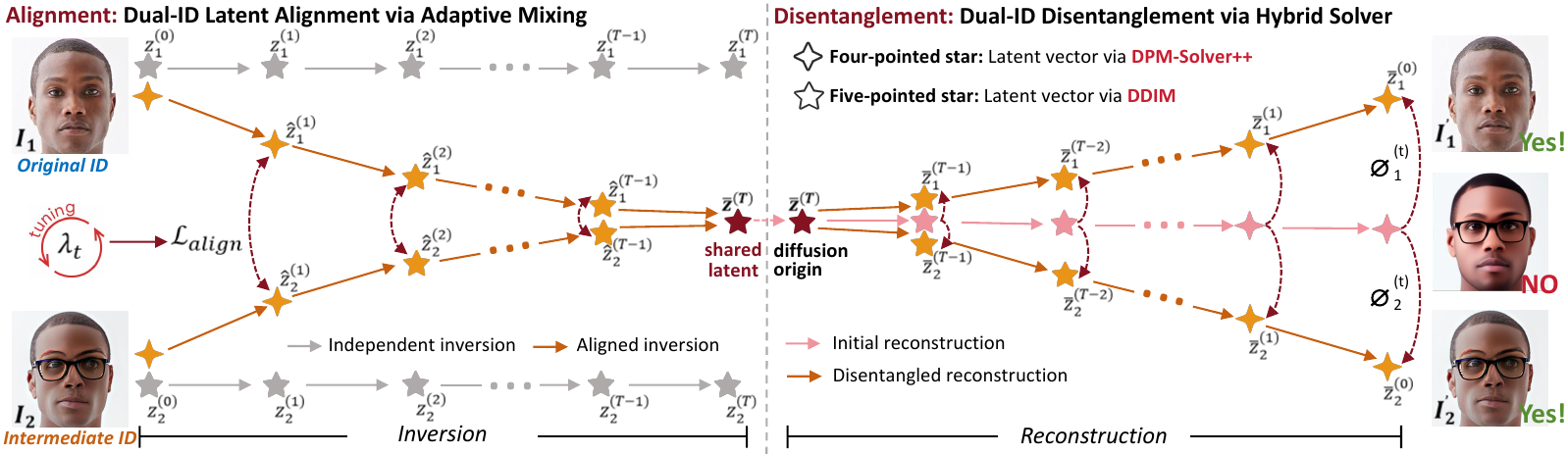}
    \caption{Dual-ID alignment (Left) and disentanglement (Right).}
  \label{fig:Alignment}
\end{figure*}
\subsection{Alignment: Dual-ID Latent Alignment via Adaptive Mixing}\label{S3.1}
Linear blending the invert trajectories of the source images $I_{1}$ and $I_{2}$ by combining their latent codes $z_{1}^{(t)}$ and $z_{2}^{(t)}$ with \emph{fixed weights} leads to two critical failures: 
1) Early inversion (near $z^{(0)}$), excessively abrupt averaging of latents causes significant degradation of source-specific features; 2) Late inversion (near $z^{(T)}$), lack of adaptivity results in overlapping trajectories, leading to cross-source feature contamination.
To address \textbf{C1} based on \textbf{Observation 1}: \emph{Multi-solution and Controllability} of trajectory,
we propose Adaptive Mixing with learnable weights $\lambda _{t}$ (\cref{fig:Alignment}). The $\lambda _{t}$ dynamically balances the contributions of $\hat{z}_{1}^{(t)}$ and $\hat{z}_{2}^{(t)}$, optimized via gradient descent to minimize alignment loss (in red):
\begin{equation}
  \mathcal{L}_{align}  = \left \| \hat{z}_{1}^{(t)} - \hat{z}_{2}^{(t)} \right \| _{2}^{2} , \quad
  \lambda _{t} \gets \lambda _{t}-\eta \cdot \bigtriangledown _{\lambda _{t}} \mathcal{L}_{align}, 
  \label{eq:important}
\end{equation}
where $\eta = 0.01$ is the learning rate, $\lambda _{t}\in \left [ 0,0.5 \right ]$. During inversion, latent codes ($\hat{z}_{1}^{(t)}$ and $\hat{z}_{2}^{(t)}$) are updated independently (noise predicted separately for $I_{1}$ and $I_{2}$ ):
\begin{equation}
\begin{split}
   \hat{z}_{1}^{(t+1)} = (1-\lambda _{t})\cdot \hat{z}_{1}^{(t)} + \lambda _{t}\cdot \hat{z}_{2}^{(t)}, \;
   \hat{z}_{2}^{(t+1)} = (1-\lambda _{t})\cdot \hat{z}_{2}^{(t)} + \lambda _{t}\cdot \hat{z}_{1}^{(t)}. 
\end{split}
  \label{eq:alignment}
\end{equation}

Lower $\lambda _{t}$ initial values yield smoother trajectory alignment.
To ensure alignment of $\hat{z}_{1}^{(t)}$ and $\hat{z}_{2}^{(t)}$ in latent space at $t=T$, we enforce constrained convergence in later diffusion steps:
\begin{equation}
   \hat{z}_{1}^{(t+1)} = \hat{z}_{2}^{(t+1)} =  (\hat{z}_{1}^{(t)}+\hat{z}_{2}^{(t)} )/2, \quad
   \lambda _{t} = 0.5.
  \label{eq:important}
\end{equation}
This yields a unified initialization $\bar{z}^{(T)}$ with two smooth merging paths (free from abrupt loss variations), which preserve original features while harmonizing latent trajectories. By storing disentangled source-specific attributes along the paths, our method achieves latent space alignment, mitigates \emph{Cross-source Distribution Bias}, and retains individuality (e.g., facial identity ($I_{1}$) and glasses ($I_{2}$)).

\subsection{Disentanglement: Dual-ID Disentanglement via Hybrid Solver}\label{S3.2}
After aligning the initial latent space $\bar{z}^{(T)}$ for dual identities, we disentangle original and intermediate ID features from this shared initialization point for selective token fusion. Extending null-text optimization, we optimize distinct null-text embeddings $\left \{  \emptyset _{i}^{(t)}   \right \}_{t=1}^{T} $ per ID to minimize MSE between reconstructed latents $\bar{z}^{(t-1)}_{i}$ and aligned states $\hat{z}^{(t-1)}_{i}$:
\begin{equation}
 \mathcal{L} _{rec} = \sum_{i=1}^{2} \left \| \hat{z}^{(t-1)}_{i}-z_{t-1}(\bar{z}^{(t)}_{i},\emptyset _{i}^{(t)},\mathcal{C}_{i})\right \| _{2}^{2}.
  \label{eq:important}
\end{equation}
Unlike single image null-text optimization, our reconstruction objective $\mathcal{L} _{rec}$ integrates joint MSE losses for both $I_{1}$ and $I_{2}$ reconstructions ($i=1,2$), conditioned on embeddings $\mathcal{C}_{i}$. 
Exclusive DDIM sampling updates subsequent latents via:
$ \bar{z}^{(t-1)}_{i} = z_{t-1}(\bar{z}^{(t)}_{i},\emptyset ^{(t)}_{i},\mathcal{C}_{i})$. Under trajectory alignment constraints, this amplifies detail degradation and artifacts in the reconstructed IDs.

To resolve \textbf{C2} with \textbf{Observation 2}—\emph{DDIM: Identity over Details, DPM-Solver++: Details over Identity}—we propose a Hybrid Solver that dynamically invokes DDIM or DPM-Solver++ during $\bar{z}^{(t-1)}_{i}$ prediction to harness complementary advantages:
\begin{equation}
 \bar{z}^{(t-1)}_{i} =
 \left\{\begin{matrix}
 DPM\text{-}Solver(\bar{z}^{(t)}_{i},\emptyset ^{(t)}_{i},\mathcal{C}_{i}),t\in [s_{1} ,s_{2} ],\\
DDIM(\bar{z}^{(t)}_{i},\emptyset ^{(t)}_{i},\mathcal{C}_{i}),otherwise.
\end{matrix}\right.
  \label{eq:hybrid solver}
\end{equation}
where $s_{1}$ and $s_{2}$ denote start/end steps for DPM-Solver++ invocation. Empirical analysis (Appendix \ref{Appendix E}) reveals optimal reconstruction strategy: DDIM invocation in early steps (near $\bar{z}^{(T)}$) establishes robust ID feature preservation, while DPM-Solver++ activation in the late steps (near $\bar{z}^{(0)}$) actively repairs and enhances the textural details from DDIM-retained features. 
Our hybrid solver optimizes diffusion timestep sampling, enabling high-fidelity ID reconstruction in few steps while resolving the long-standing efficiency-fidelity trade-off in DDIM sampling. 
Furthermore, it ensures identity-detail consistent disentanglement for dual-ID, preventing \emph{Cross-source Feature Contamination}.

\textbf{Symmetry Constraint:}
To ensure feature alignment when using the same sampler at matching noise levels, the invocation of DPM-Solver++ should be symmetric across both the inversion and the reconstruction stages in \cref{fig:Alignment}.

\textbf{Timestep Continuity:}
During the inference of $\bar{z}^{(t-1)}_i$, the sampler requires a consistent sequence of timesteps corresponding to the full diffusion step length $t = T$. However, different samplers compute their timestep sequences differently (see Appendix \ref{Appendix B}). In our initial attempt, we adopt a fragmented scheduling approach, where the diffusion trajectory is partitioned into fixed intervals, with DDIM applied over $[0, s_1)$, DPM-Solver++ over $[s_1, s_2]$, etc.

As shown in \cref{fig:Background}(3), employing a 11-step reconstruction schedule (DDIM for steps 0–4; DPM-Solver++ for steps 5–10) under symmetric inversion leads to uncontrolled latent divergence at the transition boundary ($t = 4$). 
This discontinuity resulted in chromatic aberrations and significant reconstruction errors ($z^{(0)}_1 \ne \bar{z}^{(0)}_1$) in \cref{fig:Background}(3)-top.

To resolve this, we propose a global timestep pre-setting strategy that ensures temporal continuity across the entire $[0 \rightarrow T]$ trajectory. Specifically, we first pre-compute the full timestep sequences for both schedulers:
DDIM: $\left \{ \tau _{0}, \tau _{1},...\tau _{T-1} \right \}$,
DPM-Solver++: $\left \{\sigma  _{0}, \sigma  _{1},...\sigma _{T-1} \right \}$. 
we dynamically assign the timestep at each diffusion step $t$ as:
\begin{equation}
timestep =
\begin{cases}
\sigma_t, & t \in [s_1, s_2], \\
\tau_t, & \text{otherwise}.
\end{cases}
\label{eq:important}
\end{equation}

As shown in \cref{fig:Background}(3)-bottom, this unified scheduling resolves discontinuities at critical transition points (e.g., $t = 4$), enabling smooth latent evolution and high-fidelity photorealistic identity reconstruction. The improved alignment between $z^{(0)}_1$ and $\bar{z}^{(0)}_1$ validates the effectiveness of our hybrid-sampler scheduling in mitigating reconstruction artifacts.

\begin{figure*}[h]
  \centering
  \includegraphics[width=1\linewidth]{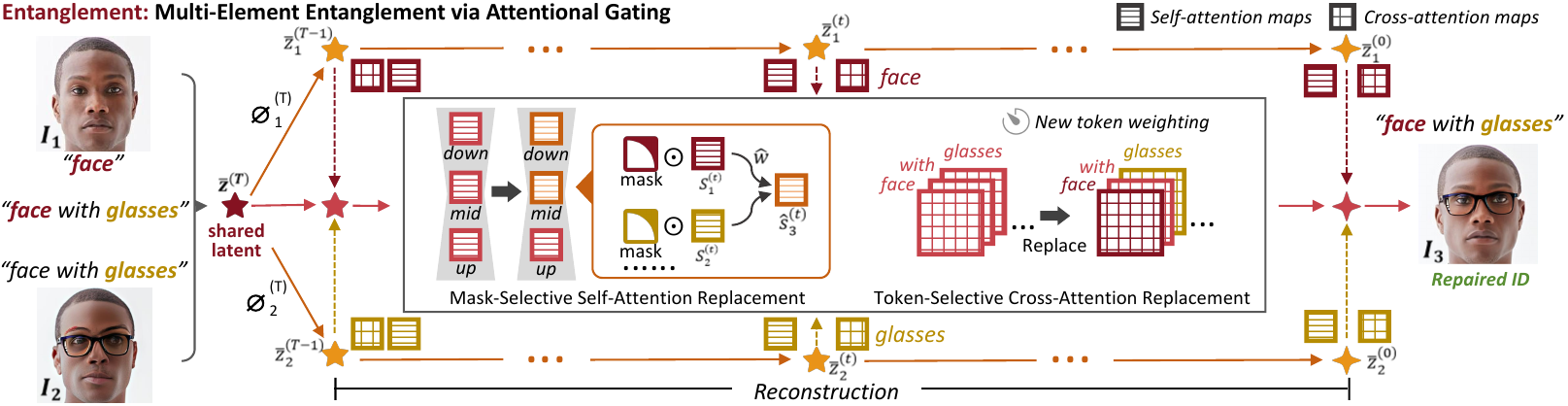}
    \caption{Multi-element entanglement: Mask/Token-selective self/cross-attention replacement.}
  \label{fig:Entanglement}
\end{figure*}
\subsection{Entanglement: Multi-Element (IP) Entanglement via Attentional Gating}\label{S3.3}
As shown in \cref{fig:Entanglement}, to address \textbf{C3}, we initiate generation from shared latent $\bar{z}^{(T)}$, reconstructing  $I_{1}$, $I_{2}$, and target $I_{3}$ under null-text constraints $\emptyset _{1}^{(t)}$ and $\emptyset _{2}^{(t)}$. During parallel diffusion, we replace attention maps of $I_{3}$ with self-attention and cross-attention maps of $I_{1}/I_{2}$ for semantic tokens. 

\textbf{Mask-Selective Self-Attention Replacement.}
In the step $t$, 
based on \textbf{Observation 3}, self-attention preserves \emph{Single-Element Structures}, 
we introduce a semantic-guided mask mechanism for self-attention feature selection. 
Specifically, spatial masks $M_{1}$ (e.g., ``face") and $M_{2}$ (e.g., ``glasses") are applied to self-attention maps $S_{1}$ (from $I_{1}$) and $S_{2}$ (from $I_{2}$) to isolate target regions:
\begin{equation}
S_{3}^{(t)} = \sum_{i=1}^{2} S_{i}^{(t)}\odot W_{i} + S_{3}^{(t)}\odot W_{3},
  \label{eq:important}
\end{equation}
For non-target regions $W_{3}$, the original self-attention map $S_{3}$ of target image $I_{3}$ is retained. The effective regions $W_{1}$ and $W_{2}$ for filtering $S_{1}$ and $S_{2}$ are defined as:
\begin{equation}
\begin{matrix}
 W_{1}=& M_{1}\cap \left ( 1-M_{2} \right ) +\hat{w}\left (M_{1}\cap  M_{2} \right ),   \\
 W_{2}=& M_{2}\cap \left ( 1-M_{1} \right ) +\left ( 1-\hat{w} \right )\cdot  \left (M_{1}\cap  M_{2} \right ), \;
 W_{3}=& 1- \left (M_{1}\cap  M_{2} \right ).
\end{matrix}
  \label{eq:entanglement}
\end{equation}
Physical interpretation: Exclusive zones of $M_{1}/M_{2}$ retain full self-attention from $S_{1}/S_{2}$. Overlapping zones apply weighted fusion with coefficient $\hat{w}$.
The entangled $S_{3}^{(t)}$ undergoes row normalization for valid attention distributions, with replacement confined to down/mid U-Net layers to preserve inter-element generability (Analysis in Appendix \ref{Appendix C}).

\textbf{Token-Selective Cross-Attention Replacement.}
At step $t$, 
based on \textbf{Observation 3}, cross-attention facilitates \emph{Multi-Element Interactions}, 
we utilize visual attention maps $A^{(t)}$ (\cref{fig:Background}(2)) to selectively entangle features between $I_{1}$ and $I_{2}$. Let $A_{1}$, $A_{2}$, $A_{3}$ denote cross-attention maps for $I_{1}$, $I_{2}$, $I_{3}$, respectively. The replacement rule can be defined as:
\begin{equation}
\begin{split}
A_{3}^{(t)}\left [ i,j \right ] = \mathbf{1}_{\left \{ i\in \mathcal{T}_{1}  \right \} }A_{1}^{(t)}\left [ i\right ]+\mathbf{1}_{\left \{ j\in \mathcal{T}_{2}  \right \} }A_{2}^{(t)}\left [ j\right ]
+\mathbf{1}_{\left \{ i\notin  \mathcal{T}_{1},j\notin  \mathcal{T}_{2}  \right \} }A_{3}^{(t)}\left [ i,j\right ].
\end{split}
\label{eq:important}
\end{equation}
where $i,j\in [0,76]$ (maximum length of text token sequence). Target token maps (e.g., ``face" from $I_{1}$, ``glasses" from $I_{2}$) are replaced throughout $t\in [0,T]$ to ensure semantic coherence.
$\mathbf{1}_{\{condition\}}$: Indicator function (1 if true, else 0);
$\mathcal{T}_{1}$,$\mathcal{T}_{2}$: Target token sets for source images;
$A^{(t)}\left [ i\right ]$: Cross-attention map for the $i$-th token. By integrating BlendDiffusion \cite{Avrahami2021BlendedDF}, our framework maintains source-specific structural priors and facilitates token-wise, context-aware interactions, enabling identity-consistent restoration without extra training.

\section{Experiment}
Experimental setup in Appendix \ref{Appendix D}, and hyperparameter sensitivity analysis in Appendix \ref{Appendix E}.
\begin{table*}[h]
\begin{floatrow}
\capbtabbox{
\renewcommand{\arraystretch}{1}{
\setlength{\tabcolsep}{0.4mm}{ 
\begin{tabular}{l|lll}
\toprule
\textbf{ID Preservation Method}                            & \textbf{ID-Sim↑}                                                                            & \textbf{CLIP-S↑}                                                                             & \textbf{I-Reward↑}                                                                          \\ \hline
\cite{rosberg2023facedancer} & 0.36                                                                                        & 25.71                                                                                        & 1.45                                                                                        \\
\cite{Ye2023IPAdapterTC}     & 0.35                                                                                        & 20.42                                                                                        & 1.02                                                                                        \\
\cite{zhao2023diffswap}      & 0.49                                                                                        & 25.85                                                                                        & 1.53                                                                                        \\
\cite{han2024face}           & 0.40                                                                                        & 26.13                                                                                        & 1.61                                                                                        \\
\cite{lin2024diffbir}        & 0.34                                                                                        & 25.43                                                                                        & 1.65                                                                                        \\
\cite{wang2025dynamicface}   & 0.63                                                                                        & 26.11                                                                                        & 1.56                                                                                        \\
\cite{baliah2025realistic}   & 0.41                                                                                        & 27.63                                                                                        & 1.68                                                                                        \\
\cite{deepfaceswap}          & 0.52                                                                                        & \underline{28.02}                                                               & 1.69                                                                                        \\
\cite{ye2025dreamid}         & \underline{0.65}                                                               & 26.11                                                                                        & \underline{1.73}                                                               \\ \hline
\textbf{EditedID (Ours)}                         & \textbf{0.73}\textcolor{red}{\tiny↑0.27} & \textbf{28.14}\textcolor{red}{\tiny↑2.43} & \textbf{1.82}\textcolor{red}{\tiny↑0.27} \\ \bottomrule
\end{tabular}
}
}
}{
    \caption{
    \textbf{Comparison with ID Preservation Methods}: ID Similarity ( \cite{deng2019arcface}), IP Preservation (CLIP-S \cite{Hessel2021CLIPScoreAR}), Human Preference (I-Reward \cite{Xu2023ImageRewardLA}).}
        \label{tab:Quantitative}
}
\capbtabbox{
\renewcommand{\arraystretch}{1}{
\setlength{\tabcolsep}{0.4mm}{ 
\begin{tabular}{l|l}
\toprule
\textbf{Multimodal Model}   & \textbf{ID-Sim↑} \\ \hline
InstructPix2Pix             & 0.37             \\
In-ContextEdit              & 0.56             \\
GPT-4o Plus                 & 0.58             \\
Flux.1 Kontext              & 0.55             \\
Doubao                      & 0.63             \\
Dreamina AI                 & 0.64             \\
BAGEL                       & 0.50             \\
Qwen-Image-Edit             & 0.52             \\ \hline
\textbf{In-Con w/ EditedID} & \textbf{0.72}\textcolor{red}{\tiny↑0.16}    \\
\textbf{Doubao w/ EditedID} & \textbf{0.75}\textcolor{red}{\tiny↑0.12}    \\ \bottomrule
\end{tabular}
}
}
}{
            \caption{\textbf{Comparison with Multimodal Editing Large Models.} In-Con denotes In-ContextEdit.}
            \label{tab:Multimodal}
 \small
}
\end{floatrow}
\end{table*}

\begin{figure*}[h]
  \centering
  \includegraphics[width=1\linewidth]{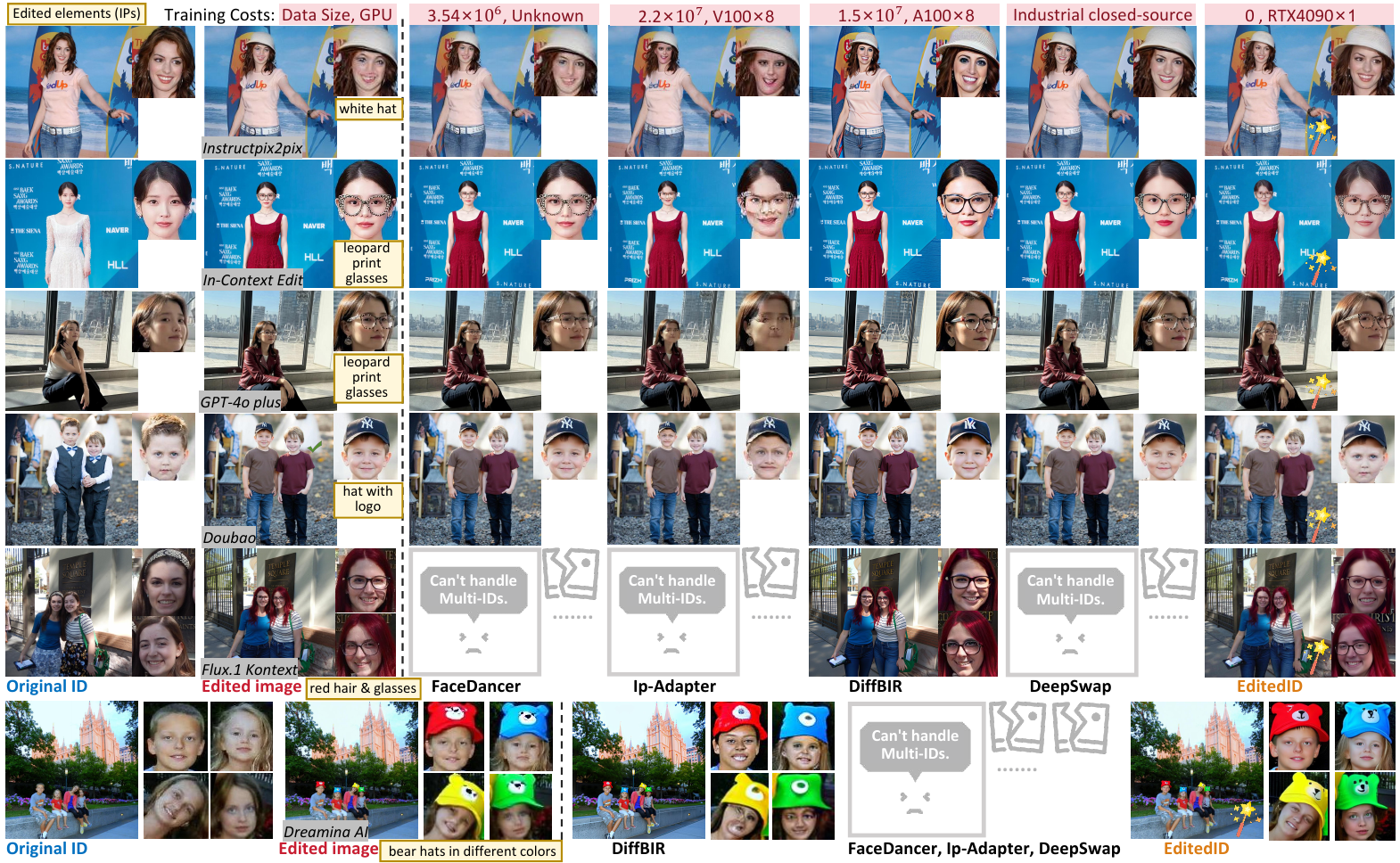}
  \caption{\textbf{Qualitative evaluation} of EditedID \emph{vs.} Industrial/Academic facial reconstruction models with training cost comparison (data size, GPU). Edited image from six  multimodal editing large models. Yellow boxes indicate edited elements (IP) that need to be preserved in the repaired ID.}
  \label{fig:Qualitative}
\end{figure*}

\textbf{Comparisons.}
As shown in \cref{fig:Qualitative} and Tab. \ref{tab:Quantitative}, we compare EditedID with SOTA methods using samples see Appendix \ref{Appendix D} for details.
The comparison spans four categories: Identity-Preserving, Identity Fusion, Blind Restoration, and Face-Swapping. We use three metrics: ID-Sim (ID similarity), CLIP-S (edited element IP preservation), and I-Reward (human-expectation compliance excluding artifacts). 
\cref{fig:Qualitative} includes challenging scenarios (details in Appendix \ref{Scenarios}): focused (single face area $> 10\%$ of image, Lines 1,2,4), multi-angle and complex lighting (Line 3), multi-person ID-specific optimization (Line 4), multi-element IP preservation (Line 5), and non-focused scenes (single face area $< 10\%$ of image, Line 6).
Key observations are summarized as follows:
\begin{itemize} [left=-10pt, itemsep=0.001em, topsep=0.001em]
    \item[ ] \includegraphics[scale=0.05]{num1.png} \emph{\textbf{EditedID resolves the trade-off between Original ID and edited Element IP preservation.}} 
    Due to \emph{Cross-source Feature Contamination} between the Original ID and Element IP, existing methods exhibit limitations: IP-Adapter often leads to \textit{facial collapse}, DiffBIR corrupts IP (e.g., transform \textit{leopard-print frames} into plain black), and FaceDancer fails to preserve \textit{high-dimensional identity features}.  By isolating \emph{Cross-source Feature Contamination} through Hybrid Solver and Attentional Gating, EditedID achieves robust and controllable identity migration while faithfully preserving all edited IP attributes, e.g., \textit{patterns}, \textit{logos}, \textit{colors}. 
    Quantitative evaluations demonstrate consistent improvements over SOTA methods, with average gains of \textbf{0.27} in \textbf{ID-Sim} and \textbf{2.43} in \textbf{CLIP-S}.
    \item[ ] \includegraphics[scale=0.05]{num2.png} \emph{\textbf{EditedID demonstrates robustness in Real-world scenarios.}} 
    Existing methods often suffer from performance degradation in real-world scenarios (e.g., profile views, challenging lighting, multi-person scenes, etc.), due to being trained or fine-tuned on facial datasets dominated by professional frontal portraits. The resulting \emph{Cross-source Distribution Bias} restricts their learnable knowledge and generalizability under real-world scenarios: IP-Adapter fails on 45° profile views, while DeepSwap loses edited IP elements and exhibits identity drift under occlusion. In contrast, EditedID mitigates \emph{Cross-source Distribution Bias} via Adaptive Mixing, achieving stable, artifact-free reconstruction in practical scenarios—yielding an average \textbf{I-Reward} improvement of \textbf{0.27} over baselines.
    \item[ ] \includegraphics[scale=0.05]{num3.png} \emph{\textbf{Existing methods struggle with Multi-IDs restoration.}} 
    Due to the inherent limitations in the application scenarios assumed by existing methods (FaceDancer, IP-Adapter, DeepSwap), they struggle to perform parallel multi-person ID restoration. In contrast, EditedID enables concurrent ID-consistent facial recovery across multiple scenarios—including multi-person focused, non-focused, ID-specific optimization, and multi-element IP preservation—through parallel identity restoration on facial patches. It also enhances original ID resolution in low-resolution defocused scenarios.

\end{itemize}
In \cref{fig:Qualitative}-top, EditedID requires neither complex fine-tuning/training nor labor-intensive data collection, achieving training-free implementation with a single GPU.
Additionally, to further evaluate the ID consistency enhancement by EditedID on multimodal editing large models, we compare it against academic \cite{brooks2023instructpix2pix,zhang2025context} and industrial state-of-the-art models \cite{flux, gpt4o,deng2025emerging, wu2025qwen,Doubao,Dreamina} in Tab. \ref{tab:Multimodal}. Clearly, existing large models exhibit facial identity degradation (ID-Sim $< 0.7$) under various challenging scenarios. This is primarily due to interference from excessive text tokens during long-instruction multi-subject editing, which dilutes the model’s focus on visual features and disrupts the retention of original identity priors, often resulting in randomly generated IDs.
Notably, EditedID offers flexible compatibility with different large models through external optimization, avoiding model-specific fine-tuning and its associated data/resource costs. After optimization, the academic model In-ContextEdit achieves a \textbf{0.16} gain in ID-Sim, while the industrial model Doubao improves by \textbf{0.12}. These results demonstrate both the effectiveness and compatibility of EditedID in enhancing ID consistency for multimodal editing large models.

\begin{figure}[h]
\centering
\begin{floatrow}
\ffigbox[0.48\textwidth]{
\includegraphics[width=\linewidth]{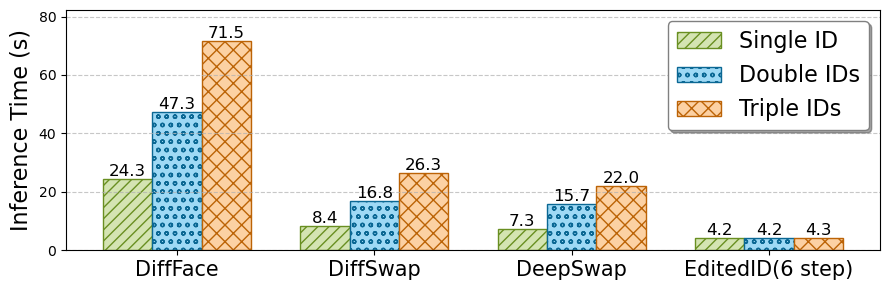}
}{\caption{Per Image Time: Single \emph{vs.} Multi IDs.}\label{fig:Time}}
\hfill
\ffigbox[0.48\textwidth]{
\includegraphics[width=\linewidth]{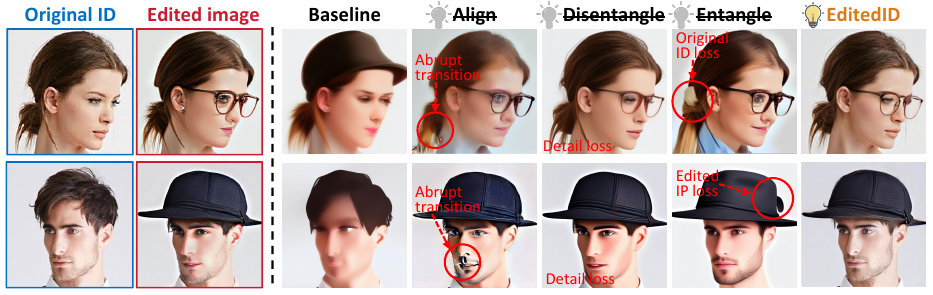}
}{\caption{Qualitative Ablation Evaluation.}\label{fig:Ablation}}
\end{floatrow}
\end{figure}

\textbf{Efficiency.}
We evaluate the inference efficiency of EditedID in both single- and multi-ID scenarios (see Fig.~\ref{fig:Time}). For the single-ID case, EditedID achieves an average reconstruction time of approximately 4.2 seconds, which is about 6$\times$ faster than the diffusion-based DiffFace (details in Appendix \ref{Inference}).
While baselines exhibit exponential time growth in multi-IDs scenarios, EditedID maintains constant inference time with parallel architecture regardless of ID count.

\textbf{Ablations.}
EditedID is evaluated through ablation (\cref{fig:Ablation} and Appendix~\ref{Quantitative Ablation}) on a baseline of null-text inverted real images with prompt-to-prompt identity fusion.
\textbf{Without alignment} introduces identity mismatches, artifacts, and abrupt transitions, highlighting the role of Adaptive Mixing in mitigating \emph{Cross-source Distribution Bias} during diffusion. 
\textbf{Without disentanglement} produces artifacts or facial distortions, validating the Hybrid Solver’s importance in balancing sampling strategies to isolate \emph{Cross-source Feature Contamination} while preserving identity and details.
\textbf{Without entanglement} causes loss of edited attributes (Elements IP e.g., hats/glasses), underscoring the necessity of Attentional Gating to maintain single-element structures and control multi-element interactions during diffusion.
In summary, EditedID is not a simple aggregation of modules, but a coherent diffusion optimization framework. Its effectiveness stems from principled improvements—Adaptive Mixing, Hybrid Solver, Attentional Gating—that optimize the pretrained diffusion process without introducing unnecessary complexity. Ablation studies confirm that each technique provides an essential refinement to diffusion dynamics, enabling stable and interpretable identity reconstruction.
Essentially, EditedID is a stable, general diffusion manipulation mechanism.

\section{Conclusion}
By reutilizing diffusion dynamics for cross-source feature fusion, we propose EditedID—a training-free and resource-efficient framework for ID-consistent face restoration. We reveal the \textbf{limitations of multimodal large models} (requiring extensive data/computation/human efforts) in preserving ID consistency; \textbf{plugging in EditedID} significantly enhances ID-consistent capability of  multimodal editing large models, demonstrating the \textbf{real-world applicability and effectiveness} of EditedID in addressing the ID consistency challenge. 
Furthermore, with its high ID consistency, EditedID serves as a pre-edit/post-edit facial \textbf{dataset calibration method} to significantly expand existing facial datasets—a single facial sample can yield multiple edited versions, thereby alleviating a \textbf{long-standing limitation} in facial ID consistency for multimodal large models—data scarcity and confidentiality.
Overall, by revealing \textbf{new insights} into diffusion Trajectories, Samplers, and Attention mechanisms, we believe EditedID will underpin future \textbf{broader applications and academic research} in multimodal fusion, deformation-resistant editing, and fine-grained attribute preservation.

\section*{Ethics Statement}

This work studies identity-consistent facial restoration and editing techniques. While the proposed method is designed for benign applications such as image restoration, creative content generation, and research purposes, we acknowledge that such technologies could be misused (e.g., for impersonation or privacy invasion).
To mitigate such risks, our work does not provide tools for identity acquisition or unauthorized identity manipulation. The method operates only on user-provided images and does not incorporate any identity recognition or retrieval components. All experiments are conducted on publicly available datasets or data with appropriate usage permissions.
We emphasize that this research is intended to advance controllable and responsible image generation. We encourage future users and developers to comply with applicable laws and ethical guidelines, obtain proper consent when handling personal data, and avoid malicious applications.

\section*{Acknowledgments}
This work is supported by the National Natural Science Foundation of China under Grants (T2541022, 62361166629), and the Key Research and Development Project of Hubei Province (2022BAD175, 2022BCA009). It is also partially supported by the WHU–Kingsoft Joint Lab. Additional support was provided by the National Natural Science Foundation of China (Grant No. 62571380).

\bibliography{iclr2026_conference}

\begin{thebibliography}{48}
\providecommand{\natexlab}[1]{#1}
\providecommand{\url}[1]{\texttt{#1}}
\expandafter\ifx\csname urlstyle\endcsname\relax
  \providecommand{\doi}[1]{doi: #1}\else
  \providecommand{\doi}{doi: \begingroup \urlstyle{rm}\Url}\fi

\bibitem[Avrahami et~al.(2022)Avrahami, Lischinski, and Fried]{Avrahami2021BlendedDF}
Omri Avrahami, Dani Lischinski, and Ohad Fried.
\newblock Blended diffusion for text-driven editing of natural images.
\newblock \emph{CVPR}, pp.\  18187--18197, 2022.

\bibitem[Baliah et~al.(2025)Baliah, Lin, Liao, Liang, and Khan]{baliah2025realistic}
Sanoojan Baliah, Qinliang Lin, Shengcai Liao, Xiaodan Liang, and Muhammad~Haris Khan.
\newblock Realistic and efficient face swapping: A unified approach with diffusion models.
\newblock In \emph{WACV}, pp.\  1062--1071, 2025.

\bibitem[{Black Forest Labs}(2025)]{flux}
{Black Forest Labs}.
\newblock Flux.1 kontext.
\newblock \url{https://www.flux1-kontext.io/}, 2025.
\newblock Accessed: 2025-5-30.

\bibitem[Brooks et~al.(2023)Brooks, Holynski, and Efros]{brooks2023instructpix2pix}
Tim Brooks, Aleksander Holynski, and Alexei~A Efros.
\newblock Instructpix2pix: Learning to follow image editing instructions.
\newblock In \emph{CVPR}, pp.\  18392--18402, 2023.

\bibitem[{ByteDance}(2025{\natexlab{a}})]{Doubao}
{ByteDance}.
\newblock Doubao.
\newblock \url{https://www.doubao.com/}, 2025{\natexlab{a}}.

\bibitem[{ByteDance}(2025{\natexlab{b}})]{Dreamina}
{ByteDance}.
\newblock Dreamina ai.
\newblock \url{https://jimeng.jianying.com/}, 2025{\natexlab{b}}.

\bibitem[Chen et~al.(2024)Chen, Ye, Qi, and Du]{10328884}
Cuiqun Chen, Mang Ye, Meibin Qi, and Bo~Du.
\newblock Sketchtrans: Disentangled prototype learning with transformer for sketch-photo recognition.
\newblock \emph{IEEE Transactions on Pattern Analysis and Machine Intelligence}, 46\penalty0 (5):\penalty0 2950--2964, 2024.
\newblock \doi{10.1109/TPAMI.2023.3337005}.

\bibitem[{DeepFaceSwap AI}(2025)]{deepfaceswap}
{DeepFaceSwap AI}.
\newblock Deepfaceswap.
\newblock \url{https://deepfaceswap.ai/}, 2025.

\bibitem[Deng et~al.(2025)Deng, Zhu, Li, Gou, Li, Wang, Zhong, Yu, Nie, Song, et~al.]{deng2025emerging}
Chaorui Deng, Deyao Zhu, Kunchang Li, Chenhui Gou, Feng Li, Zeyu Wang, Shu Zhong, Weihao Yu, Xiaonan Nie, Ziang Song, et~al.
\newblock Emerging properties in unified multimodal pretraining.
\newblock \emph{arXiv preprint arXiv:2505.14683}, 2025.

\bibitem[Deng et~al.(2019)Deng, Guo, Xue, and Zafeiriou]{deng2019arcface}
Jiankang Deng, Jia Guo, Niannan Xue, and Stefanos Zafeiriou.
\newblock Arcface: Additive angular margin loss for deep face recognition.
\newblock In \emph{CVPR}, pp.\  4690--4699, 2019.

\bibitem[Dong \& Ye(2025)Dong and Ye]{dong2025pose}
Yuran Dong and Mang Ye.
\newblock Pose-star: Anatomy-aware editing for open-world fashion images.
\newblock In \emph{ICCV}, pp.\  15822--15831, 2025.

\bibitem[Fardo et~al.(2016)Fardo, Conforto, de~Oliveira, and Rodrigues]{Fardo2016AFE}
Fernando~A. Fardo, Victor~H. Conforto, Francisco~C. de~Oliveira, and Paulo S{\'e}rgio~Silva Rodrigues.
\newblock A formal evaluation of psnr as quality measurement parameter for image segmentation algorithms.
\newblock \emph{ArXiv}, abs/1605.07116, 2016.

\bibitem[Guo et~al.(2024)Guo, Guo, Zha, Zhang, Li, Dai, Xia, and Li]{Guo2024MambaIRv2AS}
Hang Guo, Yong Guo, Yaohua Zha, Yulun Zhang, Wenbo Li, Tao Dai, Shu-Tao Xia, and Yawei Li.
\newblock Mambairv2: Attentive state space restoration.
\newblock \emph{ArXiv}, abs/2411.15269, 2024.

\bibitem[Han et~al.(2024)Han, Zhu, He, Chen, Ge, Li, Li, Zhang, Wang, and Liu]{han2024face}
Yue Han, Junwei Zhu, Keke He, Xu~Chen, Yanhao Ge, Wei Li, Xiangtai Li, Jiangning Zhang, Chengjie Wang, and Yong Liu.
\newblock Face-adapter for pre-trained diffusion models with fine-grained id and attribute control.
\newblock In \emph{ECCV}, pp.\  20--36, 2024.

\bibitem[He et~al.(2016)He, Zhang, Ren, and Sun]{He2015DeepRL}
Kaiming He, X.~Zhang, Shaoqing Ren, and Jian Sun.
\newblock Deep residual learning for image recognition.
\newblock \emph{CVPR}, pp.\  770--778, 2016.

\bibitem[Hertz et~al.(2022)Hertz, Mokady, Tenenbaum, Aberman, Pritch, and Cohen-Or]{Hertz2022PrompttoPromptIE}
Amir Hertz, Ron Mokady, Jay~M. Tenenbaum, Kfir Aberman, Yael Pritch, and Daniel Cohen-Or.
\newblock Prompt-to-prompt image editing with cross attention control.
\newblock \emph{ArXiv}, abs/2208.01626, 2022.

\bibitem[Hessel et~al.(2021)Hessel, Holtzman, Forbes, Bras, and Choi]{Hessel2021CLIPScoreAR}
Jack Hessel, Ari Holtzman, Maxwell Forbes, Ronan~Le Bras, and Yejin Choi.
\newblock Clipscore: A reference-free evaluation metric for image captioning.
\newblock \emph{ArXiv}, abs/2104.08718, 2021.

\bibitem[Ho et~al.(2020)Ho, Jain, and Abbeel]{ho2020denoising}
Jonathan Ho, Ajay Jain, and Pieter Abbeel.
\newblock Denoising diffusion probabilistic models.
\newblock \emph{Advances in neural information processing systems}, 33:\penalty0 6840--6851, 2020.

\bibitem[Hu et~al.(2025)Hu, Jin, Yao, and Lu]{Hu2025UniversalIR}
Jiakui Hu, Lujia Jin, Zhengjian Yao, and Yanye Lu.
\newblock Universal image restoration pre-training via degradation classification.
\newblock \emph{ArXiv}, abs/2501.15510, 2025.

\bibitem[Ito et~al.(2025)Ito, Viriyavisuthisakul, Kawamoto, and Kera]{ito2025undertrained}
Ru~Ito, Supatta Viriyavisuthisakul, Kazuhiko Kawamoto, and Hiroshi Kera.
\newblock Undertrained image reconstruction for realistic degradation in blind image super-resolution.
\newblock \emph{arXiv preprint arXiv:2503.02767}, 2025.

\bibitem[Li et~al.(2023)Li, Saito, Simon, Lombardi, Li, and Saragih]{li2023megane}
Junxuan Li, Shunsuke Saito, Tomas Simon, Stephen Lombardi, Hongdong Li, and Jason Saragih.
\newblock Megane: Morphable eyeglass and avatar network.
\newblock In \emph{Proceedings of the IEEE/CVF Conference on Computer Vision and Pattern Recognition}, pp.\  12769--12779, 2023.

\bibitem[Li et~al.(2024)Li, Lv, Yu, Liu, Lin, and Zhang]{Li2024}
Zonglin Li, Xiaoqian Lv, Wei Yu, Qinglin Liu, Jingbo Lin, and Shengping Zhang.
\newblock Face shape transfer via semantic warping.
\newblock \emph{Visual Intelligence}, 2, 2024.
\newblock \doi{10.1007/s44267-024-00058-7}.

\bibitem[Lin et~al.(2024)Lin, He, Chen, Lyu, Dai, Yu, Qiao, Ouyang, and Dong]{lin2024diffbir}
Xinqi Lin, Jingwen He, Ziyan Chen, Zhaoyang Lyu, Bo~Dai, Fanghua Yu, Yu~Qiao, Wanli Ouyang, and Chao Dong.
\newblock Diffbir: Toward blind image restoration with generative diffusion prior.
\newblock In \emph{ECCV}, pp.\  430--448, 2024.

\bibitem[Liu et~al.(2024)Liu, Ye, and Du]{Liu2024}
Fangyi Liu, Mang Ye, and Bo~Du.
\newblock Learning a generalizable re-identification model from unlabelled data with domain-agnostic expert.
\newblock \emph{Visual Intelligence}, 2, 2024.
\newblock \doi{10.1007/s44267-024-00062-x}.

\bibitem[Liu et~al.(2025)Liu, Han, Xing, Yin, Wang, Cheng, Liao, Wang, Fu, Han, Li, Peng, Sun, Wu, Cai, Ge, Ming, Xia, Zeng, Zhu, Jiao, Zhang, Yu, and Jiang]{Liu2025Step1XEditAP}
Shiyu Liu, Yucheng Han, Peng Xing, Fukun Yin, Rui Wang, Wei Cheng, Jiaqi Liao, Yingming Wang, Honghao Fu, Chunrui Han, Guopeng Li, Yuang Peng, Quan Sun, Jingwei Wu, Yan Cai, Zheng Ge, Ranchen Ming, Lei Xia, Xianfang Zeng, Yibo Zhu, Binxing Jiao, Xiangyu Zhang, Gang Yu, and Daxin Jiang.
\newblock Step1x-edit: A practical framework for general image editing.
\newblock \emph{ArXiv}, abs/2504.17761, 2025.
\newblock URL \url{https://api.semanticscholar.org/CorpusID:278033726}.

\bibitem[Lu et~al.(2022)Lu, Zhou, Bao, Chen, Li, and Zhu]{Lu2022DPMSolverFS}
Cheng Lu, Yuhao Zhou, Fan Bao, Jianfei Chen, Chongxuan Li, and Jun Zhu.
\newblock Dpm-solver++: Fast solver for guided sampling of diffusion probabilistic models.
\newblock \emph{ArXiv}, abs/2211.01095, 2022.

\bibitem[Mokady et~al.(2023)Mokady, Hertz, Aberman, Pritch, and Cohen-Or]{mokady2023null}
Ron Mokady, Amir Hertz, Kfir Aberman, Yael Pritch, and Daniel Cohen-Or.
\newblock Null-text inversion for editing real images using guided diffusion models.
\newblock In \emph{CVPR}, pp.\  6038--6047, 2023.

\bibitem[Nam et~al.(2024)Nam, Kim, Lee, Jin, Kim, and Chang]{Nam2024DreamMatcherAM}
Jisu Nam, Heesu Kim, Dongjae Lee, Siyoon Jin, Seungryong Kim, and Seunggyu Chang.
\newblock Dreammatcher: Appearance matching self-attention for semantically-consistent text-to-image personalization.
\newblock \emph{CVPR}, pp.\  8100--8110, 2024.

\bibitem[Ohayon et~al.(2024)Ohayon, Michaeli, and Elad]{Ohayon2024PosteriorMeanRF}
Guy Ohayon, Tomer Michaeli, and Michael Elad.
\newblock Posterior-mean rectified flow: Towards minimum mse photo-realistic image restoration.
\newblock \emph{ArXiv}, abs/2410.00418, 2024.

\bibitem[{OpenAI}(2025)]{gpt4o}
{OpenAI}.
\newblock Gpt-4o plus.
\newblock \url{https://openai.com/index/hello-gpt-4o/}, 2025.
\newblock Accessed: 2025-3-25.

\bibitem[Rosberg et~al.(2023)Rosberg, Aksoy, Alonso-Fernandez, and Englund]{rosberg2023facedancer}
Felix Rosberg, Eren~Erdal Aksoy, Fernando Alonso-Fernandez, and Cristofer Englund.
\newblock Facedancer: Pose-and occlusion-aware high fidelity face swapping.
\newblock In \emph{CVPR}, pp.\  3454--3463, 2023.

\bibitem[Song et~al.(2020)Song, Meng, and Ermon]{Song2020DenoisingDI}
Jiaming Song, Chenlin Meng, and Stefano Ermon.
\newblock Denoising diffusion implicit models.
\newblock \emph{ArXiv}, abs/2010.02502, 2020.

\bibitem[Tang et~al.(2023)Tang, Jia, Wang, Phoo, and Hariharan]{Tang2023EmergentCF}
Luming Tang, Menglin Jia, Qianqian Wang, Cheng~Perng Phoo, and Bharath Hariharan.
\newblock Emergent correspondence from image diffusion.
\newblock \emph{ArXiv}, abs/2306.03881, 2023.

\bibitem[Wang(2024)]{Wang2024FaceSV}
Feifei Wang.
\newblock Face swap via diffusion model.
\newblock \emph{ArXiv}, abs/2403.01108, 2024.

\bibitem[Wang et~al.(2025)Wang, Xu, He, Chen, Zhu, Song, Chen, Tang, and Hu]{wang2025dynamicface}
Runqi Wang, Sijie Xu, Tianyao He, Yang Chen, Wei Zhu, Dejia Song, Nemo Chen, Xu~Tang, and Yao Hu.
\newblock Dynamicface: High-quality and consistent video face swapping using composable 3d facial priors.
\newblock \emph{arXiv preprint arXiv:2501.08553}, 2025.

\bibitem[Wang \& Ye(2024)Wang and Ye]{Wang2024TexFitTF}
Tongxin Wang and Mang Ye.
\newblock Texfit: Text-driven fashion image editing with diffusion models.
\newblock In \emph{AAAI}, 2024.

\bibitem[Wu et~al.(2025)Wu, Li, Zhou, Lin, Gao, Yan, Yin, Bai, Xu, Chen, et~al.]{wu2025qwen}
Chenfei Wu, Jiahao Li, Jingren Zhou, Junyang Lin, Kaiyuan Gao, Kun Yan, Sheng-ming Yin, Shuai Bai, Xiao Xu, Yilei Chen, et~al.
\newblock Qwen-image technical report.
\newblock \emph{arXiv preprint arXiv:2508.02324}, 2025.

\bibitem[Xu et~al.(2023)Xu, Liu, Wu, Tong, Li, Ding, Tang, and Dong]{Xu2023ImageRewardLA}
Jiazheng Xu, Xiao Liu, Yuchen Wu, Yuxuan Tong, Qinkai Li, Ming Ding, Jie Tang, and Yuxiao Dong.
\newblock Imagereward: Learning and evaluating human preferences for text-to-image generation.
\newblock \emph{ArXiv}, abs/2304.05977, 2023.

\bibitem[Yang et~al.(2024{\natexlab{a}})Yang, Wu, Ren, Xie, and Zhang]{yang2024pixel}
Tao Yang, Rongyuan Wu, Peiran Ren, Xuansong Xie, and Lei Zhang.
\newblock Pixel-aware stable diffusion for realistic image super-resolution and personalized stylization.
\newblock In \emph{ECCV}, pp.\  74--91, 2024{\natexlab{a}}.

\bibitem[Yang et~al.(2024{\natexlab{b}})Yang, Jiang, Hong, Teng, Zheng, Dong, Ding, and Tang]{Yang2024InfDiTUA}
Zhuoyi Yang, Heyang Jiang, Wenyi Hong, Jiayan Teng, Wendi Zheng, Yuxiao Dong, Ming Ding, and Jie Tang.
\newblock Inf-dit: Upsampling any-resolution image with memory-efficient diffusion transformer.
\newblock \emph{ArXiv}, abs/2405.04312, 2024{\natexlab{b}}.

\bibitem[Yao et~al.(2025)Yao, Ren, Jiang, Wei, Wang, Li, and Feng]{yao2025freegraftor}
Zebin Yao, Lei Ren, Huixing Jiang, Chen Wei, Xiaojie Wang, Ruifan Li, and Fangxiang Feng.
\newblock Freegraftor: Training-free cross-image feature grafting for subject-driven text-to-image generation.
\newblock \emph{arXiv preprint arXiv:2504.15958}, 2025.

\bibitem[Ye et~al.(2025)Ye, Hua, Zhang, Li, Sun, Zhao, He, and Wu]{ye2025dreamid}
Fulong Ye, Miao Hua, Pengze Zhang, Xinghui Li, Qichao Sun, Songtao Zhao, Qian He, and Xinglong Wu.
\newblock Dreamid: High-fidelity and fast diffusion-based face swapping via triplet id group learning.
\newblock \emph{arXiv preprint arXiv:2504.14509}, 2025.

\bibitem[Ye et~al.(2023{\natexlab{a}})Ye, Zhang, Liu, Han, and Yang]{Ye2023IPAdapterTC}
Hu~Ye, Jun Zhang, Siyi Liu, Xiao Han, and Wei Yang.
\newblock Ip-adapter: Text compatible image prompt adapter for text-to-image diffusion models.
\newblock \emph{ArXiv}, abs/2308.06721, 2023{\natexlab{a}}.

\bibitem[Ye et~al.(2023{\natexlab{b}})Ye, Wu, Chen, and Du]{ye2023channel}
Mang Ye, Zesen Wu, Cuiqun Chen, and Bo~Du.
\newblock Channel augmentation for visible-infrared re-identification.
\newblock \emph{IEEE Transactions on Pattern Analysis and Machine Intelligence}, 46\penalty0 (4):\penalty0 2299--2315, 2023{\natexlab{b}}.

\bibitem[Yue \& Loy(2022)Yue and Loy]{Yue2022DifFaceBF}
Zongsheng Yue and Chen~Change Loy.
\newblock Difface: Blind face restoration with diffused error contraction.
\newblock \emph{IEEE Transactions on Pattern Analysis and Machine Intelligence}, 46:\penalty0 9991--10004, 2022.
\newblock URL \url{https://api.semanticscholar.org/CorpusID:254591838}.

\bibitem[Zhang et~al.(2025)Zhang, Xie, Lu, Yang, and Yang]{zhang2025context}
Zechuan Zhang, Ji~Xie, Yu~Lu, Zongxin Yang, and Yi~Yang.
\newblock In-context edit: Enabling instructional image editing with in-context generation in large scale diffusion transformer.
\newblock \emph{arXiv preprint arXiv:2504.20690}, 2025.

\bibitem[Zhao et~al.(2023)Zhao, Rao, Shi, Liu, Zhou, and Lu]{zhao2023diffswap}
Wenliang Zhao, Yongming Rao, Weikang Shi, Zuyan Liu, Jie Zhou, and Jiwen Lu.
\newblock Diffswap: High-fidelity and controllable face swapping via 3d-aware masked diffusion.
\newblock \emph{CVPR}, 2023.

\bibitem[Zheng et~al.(2022)Zheng, Yang, Zhang, Bao, Chen, Huang, Yuan, Chen, Zeng, and Wen]{zheng2022general}
Yinglin Zheng, Hao Yang, Ting Zhang, Jianmin Bao, Dongdong Chen, Yangyu Huang, Lu~Yuan, Dong Chen, Ming Zeng, and Fang Wen.
\newblock General facial representation learning in a visual-linguistic manner.
\newblock In \emph{CVPR}, pp.\  18697--18709, 2022.

\end{thebibliography}
\bibliographystyle{iclr2026_conference}

\newpage
\appendix
\section{Appendix}
\subsection{Related Work}
\label{Related Work}
\;\; \textbf{Identity-Preserving} approaches \cite{Yue2022DifFaceBF, Ohayon2024PosteriorMeanRF, Ye2023IPAdapterTC, ye2025dreamid} typically rely on local fine-tuning of diffusion models using edited image pairs. IP-Adapter \cite{Ye2023IPAdapterTC} has a significant disparity in training data volume. This leads to the loss of fine-grained identity features such as skin details and hair texture, rendering it ineffective for realistic facial restoration.
DreamID \cite{ye2025dreamid} adopts a triplet-based identity supervision strategy within a dual U-Net architecture. While this enables strong stylization capabilities, it suffers from high training complexity and constrained generalization due to limited and structured training data. 

\textbf{Identity Fusion} approaches \cite{yao2025freegraftor, Nam2024DreamMatcherAM, Tang2023EmergentCF} primarily target identity transfer in general scenarios. Although FreeGraftor \cite{yao2025freegraftor} and DreamMatcher \cite{Nam2024DreamMatcherAM} emphasize editable transfer rather than pure identity transfer, prioritizing flexibility in editing over strict identity fidelity.
As a result, these methods perform adequately in perceptually tolerant domains (e.g., animal or cartoon identities), but fall short in perceptually sensitive contexts such as realistic human faces. 

\textbf{Blind Restoration} methods \cite{Yue2022DifFaceBF, Ohayon2024PosteriorMeanRF, lin2024diffbir} focus on enhancing fine details in degraded facial images. Early approaches such as \cite{Yue2022DifFaceBF} attempt to bridge the low-to-high quality gap by modeling intermediate diffusion states. However, these methods emphasize restoration over generation and thus struggle to recover severely degraded faces.
More recent techniques \cite{Ohayon2024PosteriorMeanRF, lin2024diffbir} are conditioned solely on the degraded target image, they lack explicit guidance for identity preservation. This limitation similarly affects recent diffusion-based super-resolution frameworks \cite{Hu2025UniversalIR, Yang2024InfDiTUA, Guo2024MambaIRv2AS, yang2024pixel, ito2025undertrained}, which despite improved perceptual quality and remain deficient in identity-specific guidance.

\textbf{Face-Swapping} methods \cite{ye2025dreamid, wang2025dynamicface, Wang2024FaceSV, deepfaceswap} enable arbitrary face replacement in open-world scenarios. However, as the core mechanism, adapting the source identity to the edited target often results in identity feature degradation due to expression changes and facial distortions, thereby yielding low identity consistency.
While recent advances such as DreamID \cite{ye2025dreamid} achieve higher identity preservation in a wide range of poses and light conditions, they remain sensitive to input quality and fail to maintain identity fidelity when target faces are heavily degraded or distorted.

\subsection{Diffusion Samplers: DDPM, DDIM, DPM-Solver++}
\label{Appendix A}

\paragraph{DDPM} employs Markov chains. It requires all $T$ steps (typically $T=1000$), with each step necessitating a call to the noise prediction model $\varepsilon _{\theta }$, resulting in high computational cost.

\paragraph{DDIM} employs non-Markovian sampling. It accelerates the generation by skipping intermediate steps (e.g., using only $T=50$ steps), significantly improving efficiency. Its key mechanisms are:

1) The $\tilde{z} ^{(0)}$–$z^{(0)}$ Alignment: Predicts the original latent image $\tilde{z} ^{(0)}$ from the noisy latent $z^{(t)}$ at step $t$:
\begin{equation}
\tilde{z} ^{(0)}  =\frac{z ^{(t)}-\sqrt{1-\bar{\alpha }_{t}}\cdot \varepsilon _{\theta } \left ( z ^{(t)},t \right )  }{\sqrt{\bar{\alpha }_{t}}},
  \label{eq:DDIM1}
\end{equation}
where $\bar{\alpha}_t$ is the noise schedule parameter. During inversion, this prediction $\tilde{z} ^{(0)}$ is forced to align with the real image $z^{(0)}$.

2) The 1st-Order Deterministic Path: Computes the denoised sample $z ^{(k)}$ in a previous target step $k$ ($k < t$) deterministically using $z^{(t)}$ and the aligned $\tilde{z} ^{(0)}$:
\begin{equation}
z ^{(k)}  =\sqrt{\bar{\alpha }_{k}}\cdot \tilde{z} ^{(0)}+ \sqrt{1-\bar{\alpha }_{k}}\cdot \frac{\left ( z^{(t)} - \sqrt{\bar{\alpha }_{t}}\cdot \tilde{z} ^{(0)}\right ) }{\sqrt{1-\bar{\alpha }_{t}}} .
  \label{eq:DDIM2}
\end{equation}
where $\bar{\alpha}_k$ is the noise schedule in step $k$. This formula, derived from the first-order Taylor expansion (Euler method), provides a low-order approximation to solve the underlying probability-flow ODE.

\paragraph{DPM-Solver++} models the reverse process as an Ordinary Differential Equation (ODE), combined with high-order numerical methods to accelerate the solving process and reduce required sampling steps (e.g., $T=10$ steps). The key processes include: 

1) High-Order Taylor Expansion: Uses the high-order Taylor expansion function $Taylor\_Expan(\cdot)$ to approximate $z ^{(k)}$:
\begin{equation}
z ^{(k)}  \approx Taylor\_Expan\left ( z ^{(t)} , \varepsilon _{\theta } \left (  z ^{(t)},t \right ),z _{\theta } \left (  z ^{(t)},t \right )...  \right ).
  \label{eq:DPM-Solver}
\end{equation}
Approximating the ODE solution with high-order terms enables traversing from $z ^{(T)}$ to $z ^{(0)}$ in a very few steps with fast sampling.

2) Hybrid Prediction ($\varepsilon _{\theta }$/$z_{\theta }$): $\varepsilon _{\theta } \left (  z ^{(t)},t \right )$ is the noise prediction model, $z _{\theta } \left (  z ^{(t)},t \right )$ is equivalent to Eq. \ref{eq:DDIM1} (i.e., $z_{\theta } \equiv \tilde{z} ^{(0)}$). The predicted $\tilde{z} ^{(0)}$ directly participates in the denoising process.    

\subsection{Analysis of DDIM and DPM-Solver++ Characteristics}
\label{DPAnalysis}
DDIM Inversion-Reconstruction preserves ID features but loses details/texture. This is because:
\begin{itemize}[left=0pt]
\item \textbf{Deterministic Path:} During inversion, DDIM forces the predicted ID $\tilde{z}^{(0)}$ to approximate the original $z^{(0)}$ (Eq. \ref{eq:DDIM1}), thereby "anchoring" the diffusion path to the original ID image $\tilde{z}^{(0)}$ to achieve strong feature preservation.
\item \textbf{1st-Order Smoothing:} DDIM's 1st-order computation (Eq. \ref{eq:DDIM2}) uses only the current time step’s gradient for a linear approximation. This approach fails to accurately capture rapid, nonlinear changes in pixel values (i.e., high-frequency details) in the high-dimensional space, resulting in smoothed details.
\item \textbf{Error Accumulation:} The solution of the current latent vector $z^{(k)}$ depends directly on the previous latent $z^{(t)}$ (Eq. \ref{eq:DDIM2}), leading to gradual error accumulation that further degrades texture quality.
\end{itemize}

DPM-Solver++ Inversion-Reconstruction generates realistic details/texture in fewer steps but fails to preserve identity, often producing random IDs. This is because:
\begin{itemize}[left=0pt]
\item \textbf{High-Order Taylor Expansion:} DPM-Solver++ (Eq. \ref{eq:DPM-Solver}) uses high-order Taylor expansion to compute $z^{(k)}$, which incorporates the rate of change (first-order derivative), the change of that rate (second-order derivative), and higher-order information. This enables a finer and more nonlinear approximation of the data distribution and denoising process, accurately reconstructing rapid pixel variations and facilitating realistic detail and texture generation.
\item \textbf{Path Deviation:} High-order Taylor expansion acts as an extrapolation technique—predicting future values based on current and historical information. During inversion, this causes the computed path to deviate from the exact $z^{(0)}$ reconstruction path, shifting toward what the model deems "plausible", thereby losing original ID features.
\item \textbf{Prior Interference:} The hybrid prediction ($\varepsilon_{\theta}$/$z_{\theta}$) means that the $\tilde{z}^{(0)}$ predicted by $z_{\theta}$ incorporates the model’s generative prior rather than strictly adhering to the input image $z^{(0)}$. Using this $\tilde{z}^{(0)}$ during denoising accumulates bias and leads to loss of the original ID.
\end{itemize}

\subsection{Timesteps Initialization (DDIM \emph{vs.} DPM-Solver++)}
\label{Appendix B}
\paragraph{DDIM Timesteps Initialization.}
The temporal discretization strategy critically governs the sampling efficiency in diffusion models. DDIM (Denoising Diffusion Implicit Models) employs a linear schedule where time steps $\tau _{i}$ follow a deterministic uniform decay from $\tau _{max}$ to $\tau _{min}$ according to the specification in \cite{Song2020DenoisingDI}:
\begin{equation}
\tau _{i} = \tau _{max}- \frac{i}{T-1}(\tau _{max}-\tau _{min}) ,\;
i\in \left \{  0,1,...,T-1\right \} 
  \label{eq:important}
\end{equation}

with boundary parameters $\tau_{\text{max}}=0.9999$ and $\tau_{\text{min}}=0.0001$. This generates a strictly linear sequence where $\bigtriangleup \tau=\tau _{i} -\tau _{i+1}$ remains constant throughout all steps, producing a uniform coverage of the diffusion trajectory. 

\paragraph{DPM-Solver++ Timesteps Initialization.}
DPM-Solver++ fundamentally diverges through its log-uniform schedule \cite{Lu2022DPMSolverFS}, defined by exponential decay of time steps $\sigma _{i}$:
\begin{equation}
\sigma _{i} = exp(\ln_{}{\sigma _{max}} +\frac{i}{T-1}(\ln_{}{\sigma _{min}}-\ln_{}{\sigma _{max}}) ),
  \label{eq:important}
\end{equation}
parameterized with $\sigma_{\text{max}}=0.99$ and $\sigma_{\text{min}}=0.01$. This configuration yields geometrically shrinking intervals $\bigtriangleup \sigma=\sigma_{i} -\sigma_{i+1}$ that concentrate step density near $t\to 0$, aligned with regions of highest curvature in the reverse diffusion ODE. The distinct homogeneity profiles arise from the core design philosophies: DDIM's linear spacing ensures predictable convergence for moderate-step sampling, while DPM-Solver++'s exponential compression maximizes information capture during critical early denoising phases for few-step efficiency. Both methods preserve boundary consistency with $\tau _{0} =\sigma _{0} \approx 1$ (pure noise) and $\tau _{T-1} =\sigma _{T-1} \approx 0$(clean data), yet their contrasting time warping—uniform versus logarithmic—reflects divergent approaches to temporal resolution allocation.

\subsection{Self-Attention Impacts in Different Layers}
\label{Appendix C}
\begin{figure*}[h]
  \centering
  \includegraphics[width=0.9\linewidth]{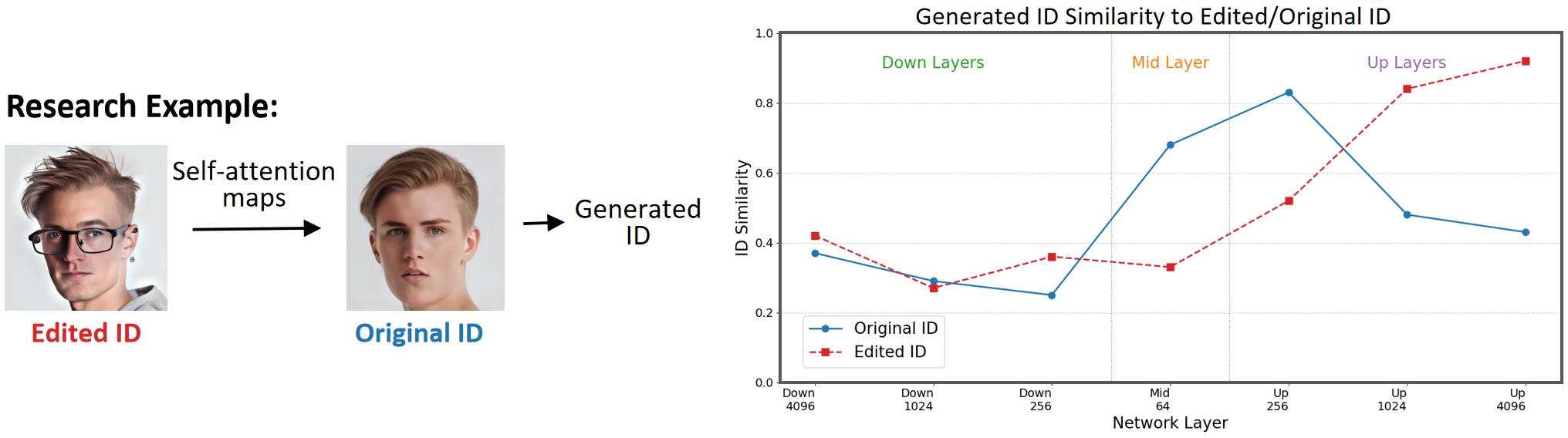}
  \caption{Layer-Specific Attention Replacement(Dual-Image Fusion).}
  \label{fig:Self-Attention}
\end{figure*}
The U-Net architecture used for noise prediction is illustrated in \cref{fig:Background}(1). Our primary focus lies on the attention layers, which comprise the downsampling blocks, the middle connection block, and the upsampling blocks. Each block contains cross-attention and self-attention maps of varying resolutions. The feature resolutions of the downsampling and upsampling blocks are 256, 1024, and 4096, while the middle block operates at a resolution of 64.

To investigate the impact on reconstruction fidelity when replacing self-attention maps from different layers within this U-Net structure, as illustrated in \cref{fig:Self-Attention}, we substitute the self-attention maps in the Original ID with corresponding maps sourced from the Edited ID during the reconstruction of the Original ID. We employ ArcFace \cite{deng2019arcface} to extract identity similarity by computing the cosine distance between the Generated ID (produced after sequentially replacing each layer) and both the Original ID and the Edited ID. Crucially, within a single generation process, only the self-attention maps belonging to a specific layer resolution are replaced, and this replacement is applied consistently throughout all diffusion steps of the generation.

The investigation of self-attention reveals two key findings:
1) Larger size attention maps capture higher-level semantics and exert greater influence on the output.
2) Replacing attention in the upsampling blocks has a more pronounced effect compared to replacements in the downsampling or middle blocks. Upsampling block replacement enhances fine-grained feature preservation but constrains generation flexibility, whereas downsampling block replacement increases generation flexibility at the cost of potential feature degradation and reduced fidelity.
These insights informed our design of the disentanglement module. To enable Mask-Selective Self-Attention Replacement to mediate interactions between source objects, based on our experimental observations, self-attention replacement is exclusively applied in the downsampling and middle blocks. Replacement in the 256-resolution layer of the upsampling block is only sporadically incorporated.

\subsection{Experimental setup}
\label{Appendix D}
\paragraph{Test Datasets.}
\begin{figure*}[h]
  \centering
  \includegraphics[width=1\linewidth]{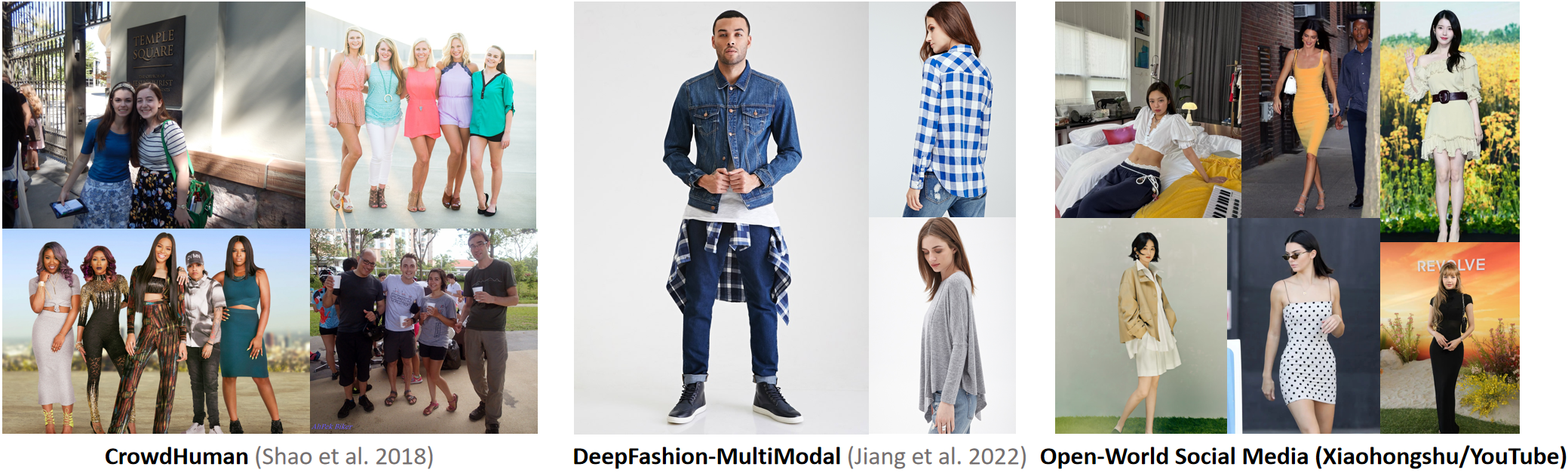}
  \caption{Composition of the test dataset.}
  \label{fig:Datasets}
\end{figure*}
Since our method does not require a training phase, large-scale training datasets are unnecessary. To evaluate practical utility in challenging real-world scenarios, our test datasets comprise: Two open-source datasets: CrowdHuman (Shao et al., 2018) and DeepFashion-MultiModal (Jiang et al., 2022); Open-World Social Media (Xiaohongshu/YouTube) samples.
As illustrated in \cref{fig:Datasets}. CrowdHuman provides crowd-sourced images (primarily from Google Search) featuring multiple individuals with various occlusions. 
DeepFashion-MultiModal contains single-person fashion images with varied facial angles, enabling robustness evaluation against pose variations.
Open-World Social Media covers celebrity images (single/multi-person) with complex environmental interference.
All 1,200 test samples are carefully curated to span challenging dimensions: multi-angle faces, complex lighting, facial occlusions, and variable ID counts. Distribution: ~30\% CrowdHuman, 35\% DeepFashion-MultiModal, 35\% Open-World Social Media.

\paragraph{Baselines.}
Our ablation baseline integrates Prompt-to-Prompt \cite{Hertz2022PrompttoPromptIE} and Null-Text Inversion \cite{mokady2023null}, reflecting the foundations of our method. First, Null-Text optimization derives latent starting points for both the original identity (ID) and the edited image. Reconstruction then proceeds from these starting points. For the target image, diffusion starts from the latent starting point of the original ID. Following Prompt-to-Prompt, cross-attention maps corresponding to target elements from both the original ID and edited image reconstructions replace the target image’s attention maps during diffusion. This iterative process yields the target image with cross-source attention guidance.
For a comprehensive evaluation, we compare four state-of-the-art categories:
Identity-Preserving Methods: Ip-Adapter, DreamID \cite{ye2025dreamid};
Identity Fusion Methods: FaceDancer \cite{rosberg2023facedancer}, FaceAdapter \cite{Ye2023IPAdapterTC};
Blind Restoration Methods: DiffBIR \cite{lin2024diffbir};
Face-Swapping Methods: DeepSwap \cite{deepfaceswap}, DiffSwap \cite{zhao2023diffswap}, CSCS \cite{wang2025dynamicface}, REFace \cite{baliah2025realistic}.

\paragraph{Implementation Details.}
Unlike existing methods requiring high-spec training/deployment hardware (e.g., 8×A100/V100), our approach eliminates training and operates on a single NVIDIA RTX 3090/4090. All experiments are performed on a single NVIDIA RTX 4090 (24GB VRAM). We use Stable Diffusion v1.4\footnote{\url{https://huggingface.co/CompVis/stable-diffusion-v1-4}} with target object masks generated by the FaRL \cite{zheng2022general} face segmentation method (supporting user customization). 
The key parameters are configured as follows: total diffusion steps = 6, initial adaptive mixing weight $\lambda _{0}$ = 0.04, DPM-Solver++ range $[s_1, s_2]$ $= [1, 3]$. The weight for the overlapping region fusion $\hat{w}$ remains user-configurable per scenario to ensure operational flexibility.

\paragraph{Evaluation Metrics.}
ID Similarity (ID-Sim) measures identity preservation between the restored image $I_{3}$ and the original identity image $I_{1}$ using ArcFace embeddings \cite{deng2019arcface}:
\begin{equation}
ID{\rm\textbf{-}}Sim=cos\left (\mathrm {f}_{ArcFace}\left ( I_{1}  \right ), \mathrm {f}_{ArcFace}\left ( I_{3}  \right )\right ),
  \label{eq:important}
\end{equation}
where $\mathrm {f}_{ArcFace}\left ( \cdot  \right ) $ denotes the ArcFace feature extractor and $cos\left ( \cdot  \right )$ computes cosine similarity.
Quantifies whether the restored image retains the original subject's identity. A threshold of $ID{\rm\textbf{-}}Sim\ge 0.7$ indicates a successful retention of identity (the same person). The values below $0.7$ signify the identity drift. 

CLIP Semantic Similarity (CLIP-S) evaluates preservation of edited attributes (e.g., glasses, accessories) in $I_{3}$. Physical Meaning: Assesses fidelity to user-specified edits. Higher values $\left ( \sim 20 \right )$ indicate better retention of edited elements \cite{Hessel2021CLIPScoreAR}.
Image Reward (I-Reward) indicates the human perceptual preference for $I_{3}$ using a learned model \cite{Xu2023ImageRewardLA}.
\begin{equation}
 I{\rm\textbf{-}}Reward=\mathcal{R}_{\theta }\left ( I_{3}  \right ) ,
  \label{eq:important}
\end{equation}
where $\mathcal{R}_{\theta }$ is a ResNet-50 \cite{He2015DeepRL} based preference predictor trained on $\sim$ 137k human judgments. Physical Meaning: Estimates visual naturalness and freedom from artifacts (e.g., distortions, blurring). Higher scores indicate better alignment with human aesthetic standards \cite{Xu2023ImageRewardLA}.

\paragraph{General Editing Model.}
To authentically demonstrate the identity (ID) consistency preservation challenges faced by state-of-the-art image editing systems, our edited images originate from four leading industrial closed-source models: GPT-4o Plus \cite{gpt4o}, Doubao \cite{Doubao}, Flux.1 Kontext \cite{flux}, and Dreamina AI \cite{Dreamina}; alongside open-sourced academic models: InstructPix2Pix \cite{brooks2023instructpix2pix} and In-Context Edit \cite{zhang2025context}. Through qualitative evaluation, we reveal significant ID consistency limitations in the outputs from these heavily-resourced models (trained with massive datasets/computational resources). We further demonstrate that integrating our training-free EditedID framework substantially enhances ID preservation capabilities across these editing platforms, thereby validating both the applicability of our approach and its effectiveness in resolving this core challenge.

\begin{figure*}[h]
  \centering
  \includegraphics[width=1\linewidth]{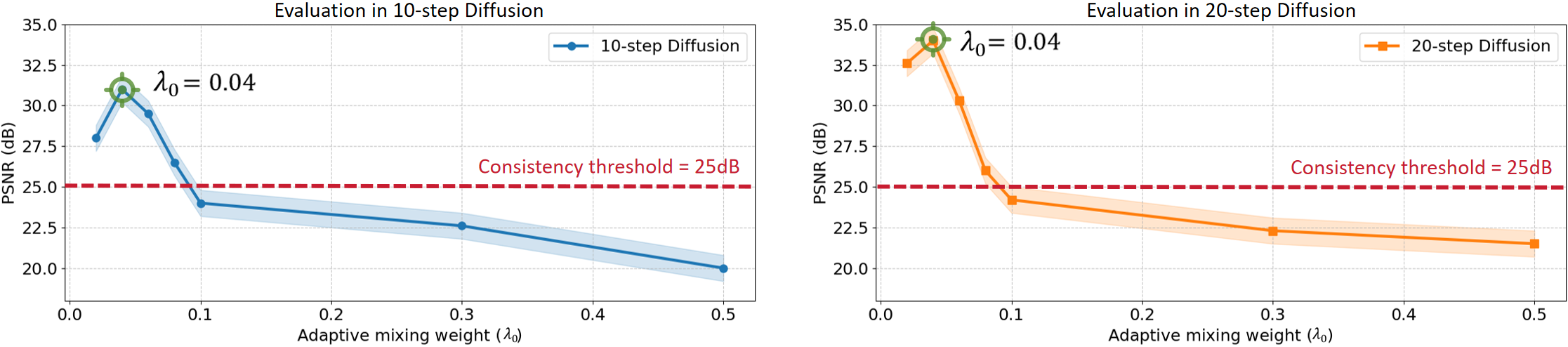}
  \caption{Quantitative evaluation of reconstruction consistency (PSNR) under different adaptive mixing weight $\lambda _{0}$ settings. The consistency threshold between reconstructed and original images is 25dB (higher PSNR indicates better quality). Results demonstrate that $\lambda _{0}$ = 0.04 achieves optimal performance across different diffusion steps.}
  \label{fig:Hyperparameter 1-1}
\end{figure*}

\subsection{Facial Corruption and Consistency Criteria Across Scenarios}
\label{Scenarios}

Based on restoration difficulty, we categorize facial corruption in images edited by multimodal large models into two levels:
High facial corruption: distorted, missing, or blended facial features making the face unrecognizable;
Low facial corruption: facial structures remain intact but exhibit inconsistent or arbitrary identity.

Our test set covers the following challenging restoration scenarios:
\begin{itemize}[left=0pt]
\item Multi-angle facial restoration: Repair of non-frontal faces (e.g., 45° or 90° profiles, overhead shots) is challenging due to the predominance of professional frontal portraits in training data.
\item Complex lighting: Requires maintaining natural lighting and shadow coherence under highly variable illumination.
\item Occluded faces: Restoration is performed around obstructions while preserving the occluding objects.
\item Multi-person ID-specific optimization: Targets identity corruption which may affect only certain subjects, requiring subject-aware restoration.
\item Multi-person multi-attribute IP preservation: Retains multiple personalized element attributes (e.g., burgundy hair and black-frame glasses) during identity recovery.
\item Focused scenes (single face area $> 10\%$ of image): Clear facial regions facilitate editing using multimodal large models, typically resulting in low corruption, though may exhibit local inconsistencies or random IDs.
\item Non-focused scenes (single face area $< 10\%$ of image): Low facial visibility editing using multimodal large models often leads to high corruption, including distortion or loss of features.
\end{itemize}


\subsection{Sensitivity}
\label{Appendix E}
The key hyperparameters are assigned to the values as follows: the initial value of the adaptive mixing weight $\lambda _{0}$ ($t=0$) of the alignment module, the DPM-Solver++ range $[s_1, s_2]$ from the disentanglement module, and the overlapping region fusion weight $\hat{w}$ from the entanglement module. The optimal hyperparameters for null-text optimization are set to \cite{mokady2023null}. Since the face-inpainting method requires strict consistency between the target element and the original element, the cross-attention and self-attention maps corresponding to the target element token are replaced throughout all diffusion steps to achieve a strictly consistent output.

\paragraph{Hyperparameter 1: adaptive mixing weight $\lambda _{0}$ ($t=0$).}

\begin{figure*}[h]
  \centering
  \includegraphics[width=1\linewidth]{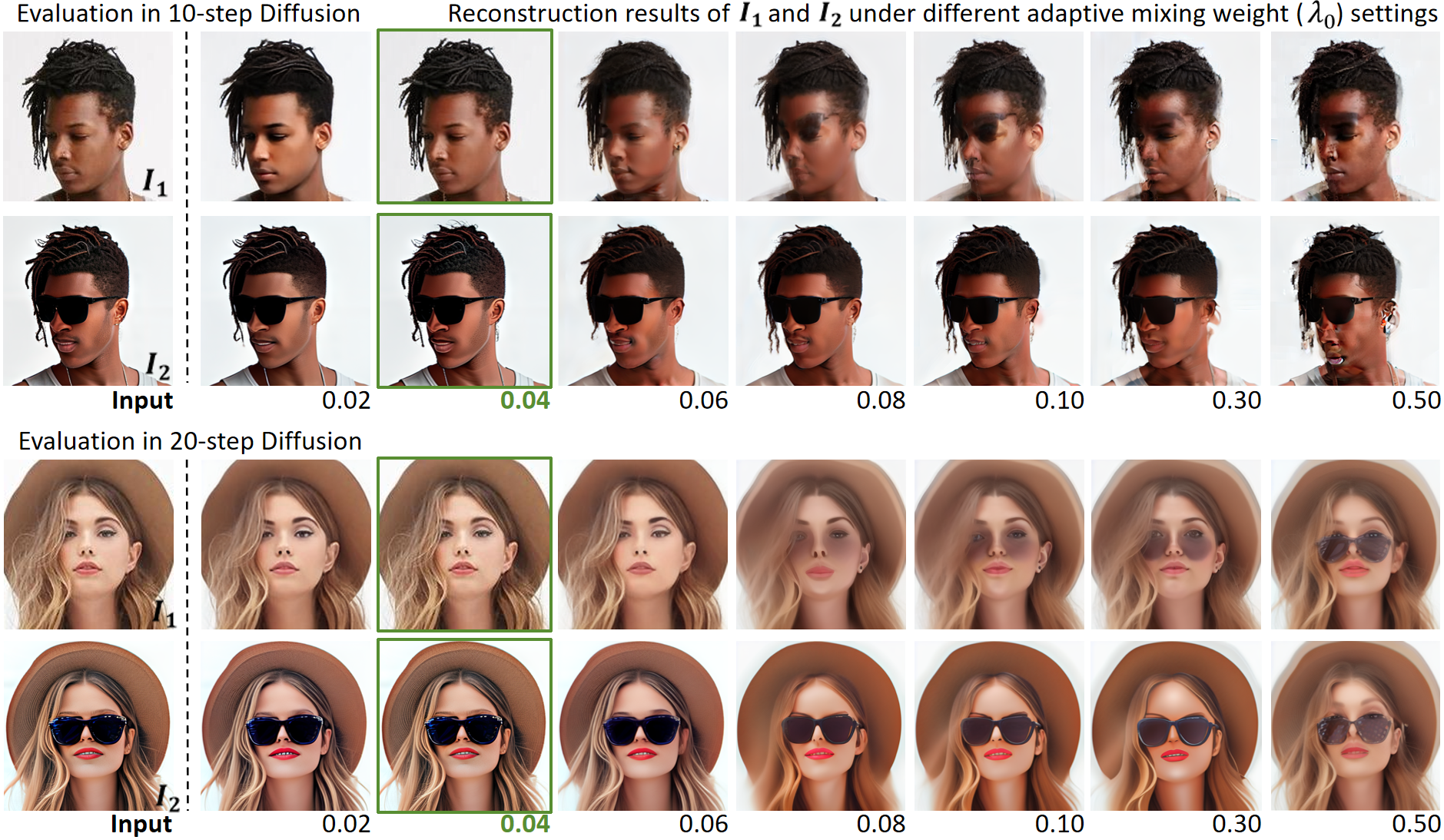}
  \caption{Qualitative evaluation of reconstruction consistency (PSNR) under different adaptive mixing weight $\lambda _{0}$ settings. Optimal performance is achieved at $\lambda _{0}$ = 0.04.}
  \label{fig:Hyperparameter 1-2}
\end{figure*}
The adaptive mixing weight $\lambda _{0}$ ($\in [0, 0.5]$), sourced from the alignment module, governs the latent space alignment between the original identity ($I_1$) and the edited image ($I_2$). This alignment is achieved through the gradient descent from the initial value of $\lambda _{0}$ at each diffusion step convergent to 0.5, where Eq. \ref{eq:alignment} confirms $I_1$-$I_2$ alignment. The initial $\lambda _{0}$ critically impacts the magnitude of the update per step: excessively large values cause disruptive updates and feature degradation, while overly small values impede smooth alignment, forcing damaging abrupt transitions near convergence. Consequently, $\lambda _{0}$'s initialization directly affects the feature consistency between the disentangled/reconstructed images ($I_1'$, $I_2'$) and the inputs ($I_1$, $I_2$).

To identify $\lambda _{0}$'s optimal initialization in diffusion steps (10-step, 20-step), we quantitatively evaluated the quality of reconstruction under varying initial $\lambda _{0}$ using PSNR. PSNR (Peak Signal-to-Noise Ratio) measures the consistency between features in source images ($I_1$, $I_2$) and reconstructions ($I_1'$, $I_2'$) from the aligned latent space. We set a PSNR threshold of 25dB, where the value $\ge 25$dB indicates the effective preservation of fine-grained features. Smaller initial $\lambda _{0}$ values mitigate excessive per-step latent updates and feature degradation during alignment, favoring gradual descent for better feature retention. Thus, our analysis focuses on the initialization of $\lambda _{0}$ within $0-0.1$. Tested values: 0.02, 0.04, 0.06, 0.08, 0.1, 0.3, 0.5.

Quantitative results for the initial $\lambda _{0}$ value (\cref{fig:Hyperparameter 1-1}) show that within the range of 0 to 0.1, the reconstruction consistency remains acceptable (PSNR $> 25$ dB) in different diffusion steps. Consistency degrades as the initial value of $\lambda _{0}$ approaches 0.5. Consequently, qualitative results (\cref{fig:Hyperparameter 1-2}) reveal that the larger initial value of $\lambda _{0}$ hinders the feature disentanglement between input images ($I_1$, $I_2$), leading to uncontrolled fusion with significant artifacts. At an initial value of $\lambda _{0}$ at 0.05, the features of $I_1$ and $I_2$ exhibit chaotic entanglement, losing original characteristics and producing random output. This validates that excessively large initial value of $\lambda _{0}$ causes destructive updates that blend multi-source features during gradient descent, resulting in mixed features in the final reconstruction. Conversely, overly small initial values (e.g., slightly above 0) cause an excessively slow gradient descent. The resulting sharp alignment shift near convergence prevents faithful reconstruction, causing detail loss (e.g., PSNR drop and facial ID detail degradation at $\lambda _{0}$=0.02, deviating from inputs $I_1$, $I_2$).

Quantitative and qualitative analysis show that 0.04 is the optimal initial value for the adaptive mixing weight $\lambda _{0}$. This also demonstrates the important role of the proposed adaptive mixing weight $\lambda _{t}$ within both the alignment and disentanglement modules.

\paragraph{Hyperparameter 2: DPM-Solver++ range $[s_1, s_2]$.}
\begin{figure*}[h]
  \centering
  \includegraphics[width=1\linewidth]{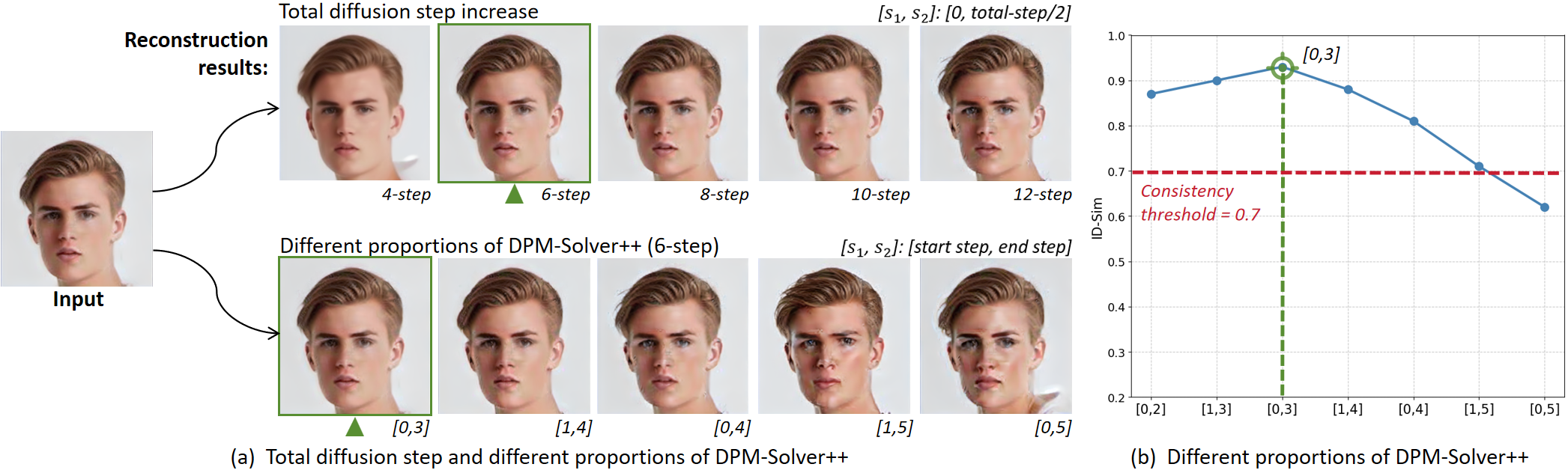}
  \caption{Impact of different diffusion steps and DPM-Solver++ proportions on the reconstruction results. $[s_1, s_2]$: starting and ending diffusion steps for DPM-Solver++ invocation. Results show that consistent input-reconstruction alignment is maintained within just 6 diffusion steps when applying DPM-Solver++ at $[0,3]$.}
  \label{fig:Hyperparameter 2}
\end{figure*}
The DPM-Solver++ invocation range $[s_1, s_2]$ ($s_1 < s_2$; $s_1, s_2 \in [0, T)$), sourced from the disentanglement module (Eq. \ref{eq:hybrid solver}), controls the step range and position where the Hybrid Solver invokes DPM-Solver++ during diffusion. Our objective is to optimize the reconstruction path via hybrid DPM-Solver++/DDIM sampling, achieving optimal fidelity to input ID features and facial details. This optimized path is then utilized in the entanglement module.

To investigate the optimal efficiency of achieving the desired reconstruction quality and the best DPM-Solver++ range, qualitative results (\cref{fig:Hyperparameter 2}(a)-top) demonstrate the reconstructions under different total diffusion steps (4, 6, 8, 10, 12), with $[s_1, s_2]$ uniformly set to $[0, T/2]$. Evidently, reconstructions achieve acceptable fidelity in just 6 steps, exhibiting high ID consistency to the input and minimal artifacts. Consequently, we set the diffusion step count for our method to 6 as the minimum viable steps, ensuring optimal efficiency with an average inference time of 4.2 seconds.

To further analyze the optimal DPM-Solver++ range $[s_1, s_2]$ under the minimum diffusion step setting (6-step), we test various start-end invocation ranges (Selected cases are shown in \cref{fig:Hyperparameter 2}(a)-bottom): $[0, 3]$, $[1, 4]$, $[0, 4]$, $[1, 5]$, $[0, 5]$. 
We observe that positioning DPM-Solver++ towards the end of the reconstruction process (closer to $\bar{z}^{(0)}$) yields superior results. This strategy leverages DDIM in the early stages to effectively preserve ID features, while employing DPM-Solver++ later reinforces ID feature details, enabling realistic facial generation in minimal steps. However, allocating an excessively large proportion of steps to DPM-Solver++ (e.g., $[0, 5]$ or $[1, 5]$) can cause ID deviation and introduce artifacts/facial blotches due to over-amplified details. Therefore, we assign DPM-Solver++ with a small number of steps in reconstruction. This optimizes detail generation and accelerates convergence without compromising original ID features. 
Based on this analysis and quantitative validation in \cref{fig:Hyperparameter 2}(b), setting $[s_1, s_2]$ to $[0, 3]$ effectively balances DDIM's ability to preserve original identity features and DPM-Solver++'s capacity for detailed refinement, thereby achieving optimal ID consistency.

This qualitative analysis of reconstruction efficiency and the optimal DPM-Solver++ range $[s_1, s_2]$ demonstrates the crucial role of the proposed hybrid solver in improving both reconstruction efficiency and quality.

\paragraph{Hyperparameter 3: overlapping region fusion weight $\hat{w}$.}
\begin{figure*}[h]
  \centering
  \includegraphics[width=1\linewidth]{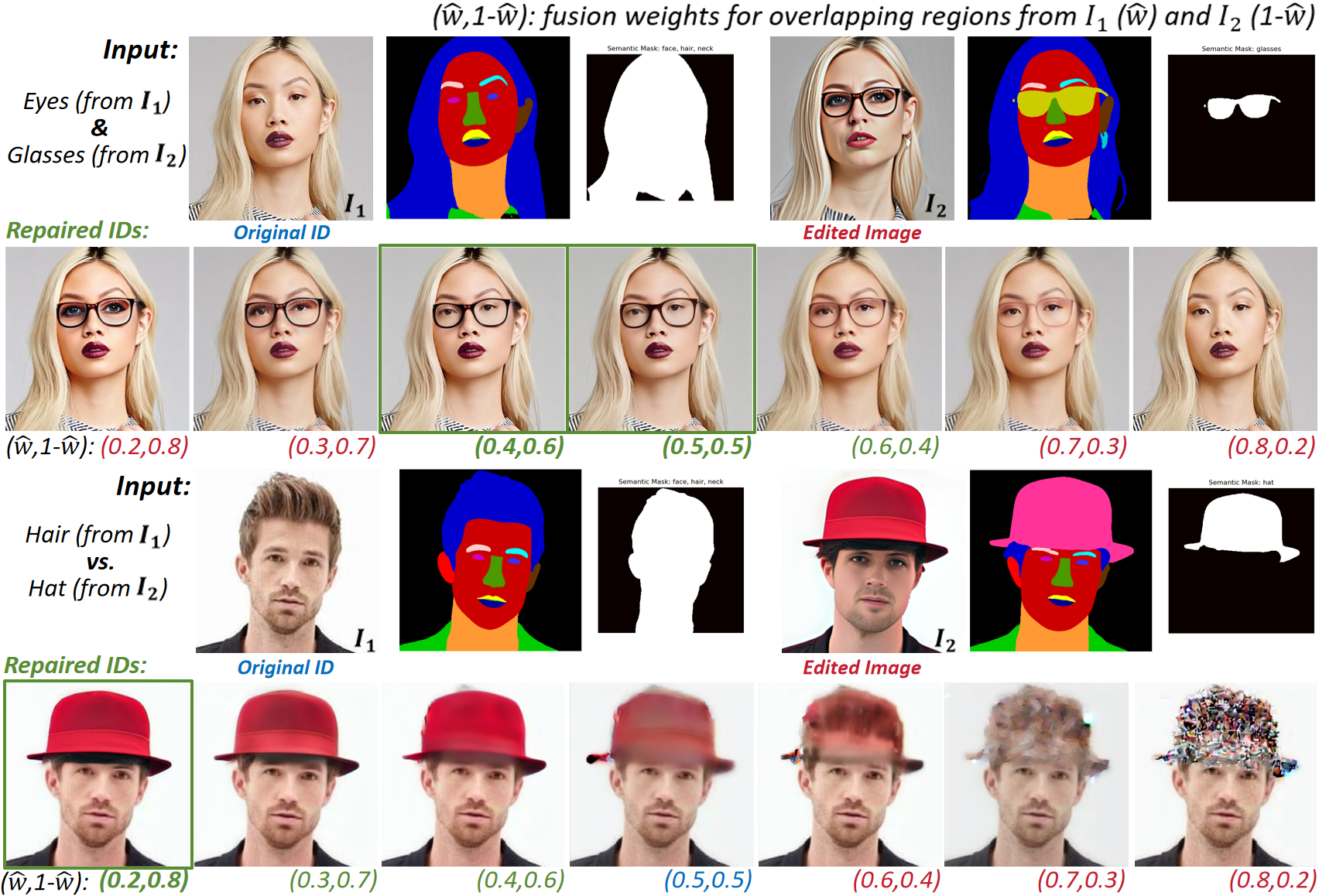}
  \caption{Inpainting results under different fusion weight $\hat{w}$ across various scenarios. $\hat{w}$: fusion weight for overlapping regions of original identity image $I_1$; 1-$\hat{w}$: fusion weight for overlapping regions of edited image $I_2$. Green color indicates physically plausible ranges (e.g., hat above hair), and red color indicates implausible ranges (e.g., hair above hat). Total diffusion steps are 6.}
  \label{fig:Hyperparameter 3}
\end{figure*}

In the entanglement module, to selectively maximize consistency between facial ID and edited elements (IP), we perform full selective replacement of self-attention for non-overlapping regions within the target object mask (extracted via the FaRL face segmentation method \cite{zheng2022general}, supporting user customization), as defined in Eq. \ref{eq:entanglement}. For overlapping regions within the target mask (e.g., transparent glasses), we aim to preserve the edited element (e.g., frames) while restoring the original ID (e.g., eyes). Thus, we employ the overlapping region fusion weight $\hat{w}$ ($\hat{w} \in [0,1]$), where $\hat{w}$ denotes the weight assigned to elements from the original ID ($I_1$) in the overlapping region, and $1-\hat{w}$ denotes the weight assigned to elements from the edited image ($I_2$).

In particular, the optimal value of $\hat{w}$ varies in different scenarios and it is user-configurable, primarily dictated by its physical interpretation in the target use case. To investigate the setting principles for $\hat{w}$ in different applications (\cref{fig:Hyperparameter 3}), we visualize the target element masks from $I_1$ and $I_2$ and show the inpainting results under various $(\hat{w}, 1-\hat{w})$ settings: (0.2, 0.8), (0.3, 0.7), (0.4, 0.6), (0.5, 0.5), (0.6, 0.4), (0.7, 0.3), (0.8, 0.2). We categorize the application scenarios into two types: (1) Co-existence: Elements from different sources need simultaneous preservation (e.g., Eyes ($I_1$) \& Glasses ($I_2$)); (2) Layer Coverage: Elements from different sources exhibit an occlusion relationship (e.g., Hair ($I_1$) \emph{vs.} Hat ($I_2$)).

Qualitative evaluation reveals the setting principles for $\hat{w}$:
\textbf{1) Co-existence Scenarios}, where optimal values approach $\hat{w} \approx 1-\hat{w}$. Specifically, when preserving eyes ($I_1$) and glasses ($I_2$) concurrently, the green-marked interval from (0.4, 0.6) to (0.6, 0.4) generally contains the best values. Since both elements are equally important, balanced $(\hat{w}, 1-\hat{w})$ weights best preserve cross-source features. In contrast, extreme values cause excessive glass retention or loss.
\textbf{2) Layer Coverage Scenarios:} Optimal values satisfy the underlying element weight ($\hat{w}$) $<$ the top element weight ($1-\hat{w}$). For hair ($I_1$, underlying) and hat ($I_2$, top), the green-marked interval (0.2, 0.8) to (0.4, 0.6) produces plausible results. The setting of $\hat{w} < 1-\hat{w}$ aligns with reality (hair under the hat). The red-marked interval from (0.6, 0.4) to (0.8, 0.2) violates diffusion priors (implausible hat-under-hair configuration), generating artifacts and blotches.

This analysis demonstrates that the proposed attentional gating is effective in selective feature fusion. The established $\hat{w}$ configuration principles provide user-actionable guidance, ensuring method flexibility in real-world applications.

\subsection{Quantitative Ablation Study}
\label{Quantitative Ablation}
Quantitative ablation results in Tab. \ref{tab:Ablation} show that removing the alignment module reduces identity consistency (-0.16 ID-Sim), disabling the disentanglement module degrades perceptual realism (-0.13 I-Reward), and eliminating the entanglement module causes the most significant drop in edited attribute preservation (-2.81 CLIP-S). These demonstrate the effectiveness of the proposed components—adaptive mixing, hybrid solver, and attentional gating—in optimizing diffusion-based facial reconstruction across three key dimensions: identity consistency, realism, and element preservation.
\begin{table}[h]
\renewcommand{\arraystretch}{0.8}
\centering
\setlength{\tabcolsep}{1mm}{
\begin{tabular}{l|lll}
\toprule
\textbf{Model}      & \multicolumn{1}{l}{\textbf{ID-Sim↑}} & \multicolumn{1}{l}{\textbf{CLIP-S↑}} & \multicolumn{1}{l}{\textbf{I-Reward↑}} \\ \hline
\textbf{Full Model} & \textbf{0.73}                        & \textbf{28.14}                       & \textbf{1.82}                          \\
w/o Alignment       & 0.57                                 & 26.92                                & 1.72                                   \\
w/o Disentanglement & 0.63                                 & 26.48                                & 1.69                                   \\
w/o Entanglement    & 0.61                                 & 25.33                                & 1.78                                   \\
Baseline            & 0.21                                 & 19.32                                & 1.03                                   \\ \bottomrule
\end{tabular}

}
  \caption{Quantitative ablation study results}
  \label{tab:Ablation}
\end{table}

\subsection{User Survey}
\label{User Survey}
To further evaluate the facial reconstruction performance of EditedID, we conducted a user study with 200 participants, including computer vision engineers, graphic designers, and lay users. This stratified sampling ensured a balanced assessment across both usability and technical stability. Participants were asked to rate the restored images on three metrics: ID Consistency, Facial Fidelity, and IP Preservation, on a 5-point scale (higher is better). The average results, shown in Tab. \ref{tab:survey}, clearly demonstrate that our method achieves competitive outcomes across all three dimensions compared to existing approaches.

\begin{table}[h]
\renewcommand{\arraystretch}{0.8}
\centering
\setlength{\tabcolsep}{1mm}{
\begin{tabular}{l|ccc}
\toprule
\textbf{Method}          & \textbf{ID Consistency↑} & \textbf{Facial Fidelity↑} & \textbf{IP Preservation↑} \\ \hline
\cite{rosberg2023facedancer}                       & 3.34\footnotesize{$\pm$0.31}                     & 3.83\footnotesize{$\pm$0.07}                      & 4.19\footnotesize{$\pm$0.03}                      \\
\cite{Ye2023IPAdapterTC}                       & 3.16\footnotesize{$\pm$0.97}                     & 2.13\footnotesize{$\pm$0.94}                      & 2.11\footnotesize{$\pm$0.33}                      \\
\cite{zhao2023diffswap}                       & 3.77\footnotesize{$\pm$0.52}                      & 3.62\footnotesize{$\pm$0.83}                      & 3.41\footnotesize{$\pm$0.43}                      \\
\cite{han2024face}                       & 3.61\footnotesize{$\pm$0.02}                      & 3.51\footnotesize{$\pm$0.40}                      & 3.48\footnotesize{$\pm$0.45}                      \\
\cite{lin2024diffbir}                       & 3.11\footnotesize{$\pm$0.90}                     & 4.02\footnotesize{$\pm$0.75}                      & 3.21\footnotesize{$\pm$0.70}                      \\
\cite{wang2025dynamicface}                       & 4.40\footnotesize{$\pm$0.33}                     & 3.55\footnotesize{$\pm$0.02}                      & 3.62\footnotesize{$\pm$0.73}                      \\
\cite{baliah2025realistic}                       & 3.32\footnotesize{$\pm$0.53}                     & 4.11\footnotesize{$\pm$0.33}                      & 3.81\footnotesize{$\pm$0.93}                      \\
\cite{deepfaceswap}                       & 4.22\footnotesize{$\pm$0.17}                     & 4.62\footnotesize{$\pm$0.93}                      & 4.01\footnotesize{$\pm$0.69}                      \\
\cite{ye2025dreamid}                       & 4.68\footnotesize{$\pm$0.10}                     & 4.78\footnotesize{$\pm$0.37}                       & 4.32\footnotesize{$\pm$0.20}                       \\ \hline
\textbf{EditedID (Ours)} & \textbf{4.85\footnotesize{$\pm$0.52} }            & \textbf{4.83\footnotesize{$\pm$0.36} }             & \textbf{4.72\footnotesize{$\pm$0.19} }             \\ \bottomrule
\end{tabular}

}
  \caption{The participants were asked to rate: (1) ID Consistency, (2) Facial Fidelity, and (3) IP Preservation. The perfect score is 5.}
  \label{tab:survey}
\end{table}

\subsection{Inference Cost Analysis}
\label{Inference}

\begin{table}[h]
\centering
\resizebox{0.77\textwidth}{!}{
\begin{tabular}{c|c|c|c}
\toprule
\textbf{Method} & \textbf{Diffusion Steps} & \textbf{Total Time (s)} & \textbf{GPU Mem (GB)} \\
\hline
Prompt-to-Prompt & 50 & $\sim$8.9 & $\sim$13.6 \\
Null-text Inversion & 50 & $\sim$56.5 & $\sim$16.2 \\
EditedID (ours) & 6 & $\sim$7.0 & $\sim$16.5 \\
\bottomrule
\end{tabular}
}
\caption{Approximate inference cost of different methods for reference.}
\end{table}

We report approximate inference cost under a reference implementation. EditedID follows the standard two-stage diffusion pipeline, consisting of inversion (alignment) and reconstruction with disentanglement/entanglement, without introducing additional inference stages.
The primary efficiency gain comes from reducing diffusion steps, while memory usage remains comparable to existing inversion-based methods. 
The reported step numbers correspond to the minimal configuration used by EditedID; we also support more flexible or increased step settings to achieve higher-quality results when desired.
All measurements are obtained under a specific experimental setup and optimized research implementation; actual runtime may vary depending on hardware and software environments.

\subsection{Attention Validation}
To verify the attention mechanism, we conducted a analysis in which attention maps from Image 2 were injected into the reconstruction process of Image 1. Specifically, we replaced (i) self-attention maps (single-element) and (ii) cross-attention maps (multi-element) across different U-Net layers, resolutions, and architectural variants. We then evaluated the impact of these substitutions by measuring structural consistency using SSIM and semantic alignment using CLIP-S between the reconstructed output and the ground-truth Image~2. This controlled setup allows us to rigorously quantify how attention at each level contributes to identity-related structural and semantic information.

The results above reveal a clear and consistent pattern:
(1) Self-attention replacements substantially improve single-element structural preservation, with the strongest effects observed in up-blocks and at higher spatial resolutions.
(2) Cross-attention replacements markedly enhance multi-element semantic alignment, and follow the same hierarchical progression across layers and resolutions.
This trend further supports the generality of our conclusions and demonstrates that the identified behaviors hold consistently across different architectural scales and variants.
\begin{table}[h]
\centering
\resizebox{0.78\textwidth}{!}{
\begin{tabular}{c|c|cc|cc}
\toprule
\multirow{2}{*}{\textbf{Layer}} & \multirow{2}{*}{\textbf{Resolution}} & \multicolumn{2}{c|}{\textbf{Replace Self-attn (SSIM)}} & \multicolumn{2}{c}{\textbf{Replace Cross-attn (CLIP-S)}} \\ \cline{3-6} 
                                &                                      & \textbf{SD1.4}  & \textbf{SDXL} & \textbf{SD1.4}   & \textbf{SDXL}  \\ \hline
\textbf{-}                      & -                                    & 0.22                       & 0.24                      & 15.22                       & 15.14                      \\
\multirow{2}{*}{\textbf{Down}}  & 32×32                                & 0.62                       & 0.67                      & 19.70                       & 21.32                      \\
                                & 16×16                                & 0.58                       & 0.65                      & 18.51                       & 20.09                      \\
\textbf{Mid}                    & 8×8                                  & 0.52                       & 0.56                      & 16.74                       & 19.21                      \\
\multirow{2}{*}{\textbf{Up}}    & 16×16                                & 0.63                       & 0.72                      & 20.37                       & 23.11                      \\
                                & 32×32                                & 0.71                       & 0.76                      & 22.98                       & 25.31                      \\ \bottomrule
\end{tabular}
}
\caption{\label{tab:widgets}Attention mechanism analysis.}
\end{table}

\subsection{Solver Stability}
We further evaluate the stability of our hybrid solver using multiple complementary metrics, including cumulative MSE, maximum transition jump, final latent L2 distance, and PSNR.
As reported in Tab. \ref{tab:Solver}, our global hybrid strategy achieves performance comparable to single-solver DDIM in terms of trajectory consistency, while yielding improved reconstruction quality. In particular, the proposed hybrid solver maintains low error accumulation and smooth transitions across solver-switching points, indicating stable inversion–reconstruction trajectories.
Overall, these results confirm that the hybrid design preserves stability while benefiting from the complementary strengths of DDIM and DPM-Solver++, leading to consistent and high-fidelity reconstruction.
\begin{table}[h]
\centering
\resizebox{\textwidth}{!}{
\begin{tabular}{c|cccc}
\toprule
\textbf{Method}               & \textbf{Cumulative MSE ($\times 10^{-3})$↓} & \textbf{maxJump ($\times 10^{-3}$)↓} & \textbf{latent L2 ($\times 10^{-2}$)↓} & \textbf{PSNR (dB)↑} \\ \hline
\textbf{DDIM-only}            & \underline{18.5 ± 3.2}                       & \textbf{1.9 ± 0.6}        & \textbf{1.7 ± 0.4}                & \underline{28.6 ± 0.9}          \\
\textbf{DPM-only}             & 24.2 ± 4.5                       & 3.7 ± 0.9                 & 2.5 ± 0.6                         & 27.2 ± 0.8          \\
\textbf{Hybrid-fragmented}    & 47.1 ± 5.0                       & 8.9 ± 2.2                 & 4.3 ± 0.8                         & 20.1 ± 1.1          \\
\textbf{Hybrid-global (ours)} & \textbf{17.8 ± 2.6}              & \underline{2.1 ± 0.5}                 & \underline{1.8 ± 0.3}                         & \textbf{29.0 ± 0.9} \\ \bottomrule
\end{tabular}
}
\caption{\label{tab:Solver}Stability of the hybrid solver.}
\end{table}

\subsection{Impact of DiffBIR Preprocessing}
We further study the influence of DiffBIR when used as a preprocessing step before EditedID. Tab. \ref{tab:DiffBIR} reports quantitative results under clean and degraded inputs.
Applying DiffBIR alone on degraded images leads to a noticeable drop in identity similarity, while combining DiffBIR with EditedID restores ID-Sim close to the clean-input setting. This suggests that EditedID can effectively compensate for identity distortions introduced by restoration models.
Qualitatively, this robustness is mainly attributed to (i) reconstructing the target identity along an independent inversion–reconstruction trajectory, (ii) selectively transferring only edited elements from DiffBIR outputs while discarding distorted identity information, and (iii) leveraging the hybrid solver to refine local texture details.
Overall, these results indicate that EditedID does not simply inherit artifacts from preprocessing models, but can mitigate identity degradation and improve visual quality. 
\begin{table}[h]
\centering
\resizebox{0.65\textwidth}{!}{
\begin{tabular}{c|cc}
\toprule
\textbf{Method}                         & \textbf{ID-Sim} & \textbf{ImageReward} \\ \hline
Clean → EditedID (w/o DiffBIR) & 0.74 ± 0.03     & 1.82 ± 0.05          \\
Degraded → DiffBIR only        & 0.36 ± 0.07     & 1.65 ± 0.08          \\
Degraded → DiffBIR → EditedID  & 0.71 ± 0.04     & 1.78 ± 0.05          \\ \bottomrule
\end{tabular}
}
\caption{\label{tab:DiffBIR}Impact of DiffBIR preprocessing on identity preservation and image quality.}
\end{table}

\subsection{Multi-Region Editing}
We report an ablation study on Attentional Gating under increasingly complex editing settings (single, dual, and three attributes). Results are summarized in Tab. \ref{tab:Multi-Region}.
Without gating, performance degrades notably as the number of edited regions increases. Introducing Attentional Gating consistently improves both element preservation (measured by CLIP-S) and identity consistency (ID-Sim), with larger gains observed in more complex multi-attribute scenarios.
These results suggest that Attentional Gating helps reduce cross-region interference during fusion, leading to more stable element control and identity retention when multiple edits are applied simultaneously.

\begin{table}[h]
\centering
\resizebox{0.85\textwidth}{!}{
\begin{tabular}{c|c|cc}
\toprule
\textbf{Editing Attribute}       & \textbf{EditedID Architecture} & \textbf{IP (CLIP-S$\ge$0.75)} & \textbf{ID (ID-Sim$\ge$0.70)} \\ \hline
\multirow{2}{*}{Single} & full                           & 92\%                                   & 94\%                                  \\
                                 & w/o gating                     & 78\%                                   & 74\%                                  \\ \hline
\multirow{2}{*}{Dual}   & full                           & 87\%                                   & 85\%                                  \\
                                 & w/o gating                     & 69\%                                   & 66\%                                  \\ \hline
\multirow{2}{*}{Three}  & full                           & 84\%                                   & 81\%                                  \\
                                 & w/o gating                     & 58\%                                   & 62\%                                  \\ \bottomrule
\end{tabular}
}
\caption{\label{tab:Multi-Region}Effect of attentional gating in multi-region editing.}
\end{table}

\subsection{Multi-Person Processing}

Multi-person editing in EditedID is implemented as an external engineering pipeline rather than an intrinsic model component. For this reason, it is not included in the main framework description and is summarized here for completeness.
Specifically, a face detector is first applied to locate all individuals in the input image. User instructions, together with positional correspondences, are used to associate edited elements with their respective identities. EditedID is then independently applied to each detected face in parallel to restore identity consistency. Finally, the processed faces are composited back into the original image according to their spatial locations, and an inpainting model is used to smooth interaction boundaries for natural transitions.
We will further update an easy-to-use version of this pipeline in the released codebase to improve practical usability.

\subsection{Limitations and Discussion}
\label{Appendix F}
Our method enables customizable facial restoration in edited portraits by manipulating diffusion trajectories across multiple images, with hard-constrained source identity features ensuring strong facial consistency with the original subject. While EditedID supports parameter-tunable restoration, its training-free nature introduces a usability barrier, limiting accessibility for non-expert users. Although the method performs reliably across a wide range of open-world fashion portrait edits, certain outlier cases may still lead to inconsistent results due to suboptimal hyperparameter configurations. 

To enhance the generalization and user-friendliness of EditedID in practical applications, future work under sufficient computational resources may focus on two directions: integrating the framework into multimodal large models or training it on real-world datasets to improve accessibility. Another promising direction is to adopt EditedID as a data generation tool. Inspired by InstructPix2Pix\cite{brooks2023instructpix2pix}, which uses generated data from Prompt-to-Prompt\cite{Hertz2022PrompttoPromptIE} to train instruction-following stable diffusion models, EditedID can help alleviate the persistent challenge of multimodal large models in facial ID consistency—a critical barrier in real-world applications due to the scarcity and confidentiality of facial data, as well as the difficulty in acquiring aligned pre- and post-edit pairs.
With its high ID consistency, EditedID can serve as a calibration model: real face images can be edited and then corrected using EditedID to form a hybrid dataset (pre-edit: real/synthetic; post-edit: generated). Such mixed data improves real-world applicability. Moreover, the diversity of editing instructions enables extensive sample expansion—a single facial sample can yield multiple edited versions, greatly augmenting dataset size and diversity.
The resulting large-scale calibrated dataset can circumvent current limitations on facial data usage and be employed to train identity-preserving models or fine-tune multimodal editing models, thereby enhancing ID retention capabilities. 
We believe that EditedID holds significant value both as an effective ID-consistent facial reconstruction framework and as a data calibration tool.
We believe that, with sufficient computational resources in the future, this method will yield more valuable research findings and deliver unexpected surprises in its applications.

\end{document}